\documentclass[10pt,twocolumn,letterpaper]{article}

\usepackage{cvpr}
\usepackage{times}
\usepackage{amsmath}
\usepackage{amssymb}
\usepackage{graphicx}
\usepackage{color}
\usepackage{subfig}
\usepackage{algorithm}
\usepackage{algorithmic}
\usepackage{multirow}

\usepackage[pagebackref=true,breaklinks=true,letterpaper=true,colorlinks,bookmarks=false]{hyperref}
\newtheorem{theorem}{Theorem}

\cvprfinalcopy % *** Uncomment this line for the final submission

\def\cvprPaperID{927} % *** Enter the CVPR Paper ID here

\ifcvprfinal\fi

\begin{document}

\title{Low-Rank Representation over the Manifold of Curves}

% \author{Stephen Tierney$^1$ and Junbin Gao\\
% School of Computing and Mathematics\\
% Charles Sturt University\\
% Bathurst, NSW 2795, Australia\\
% {\tt\small \{stierney, jbgao\}@csu.edu.au}
% \and
% Yi Guo\\
% Digital Productivity \& Services Flagship\\
% CSIRO\\
% North Ryde, NSW 2113, Australia\\
% {\tt\small yi.guo@csiro.au}
% \and
% Zhengwu Zhang\\
% Department of Statistics\\
% Statistical and Applied Mathematical Sciences Institute\\
% P.O. Box 14006, Research Triangle Park, NC 27709-4006, USA\\
% {\tt\small zhengwustat@gmail.com}
% }

\author{Stephen Tierney$^1$, Junbin Gao$^2$, Yi Guo$^3$ and Zhengwu Zhang$^4$\\
{\small $^1$School of Computing and Mathematics,
Charles Sturt University,
Bathurst, NSW 2795, Australia}\\
{\small $^2$Business Analytics Discipline, The University of Sydney Business School, Camperdown NSW 2006, Australia}\\
{\small $^3$Digital Productivity \& Services Flagship,
CSIRO, North Ryde, NSW 2113, Australia} \\
{\small $^4$Department of Statistics,
Statistical and Applied Mathematical Sciences Institute, NC 27709-4006, USA}\\
{\tt\small stierney@csu.edu.au; junbin.gao@sydney.edu.au; yi.guo@csiro.au; zhengwustat@gmail.com}
}

\maketitle

\begin{abstract}
In machine learning it is common to interpret each data point as a vector in Euclidean space. However the data may actually be functional i.e.\ each data point is a function of some variable such as time and the function is discretely sampled. The naive treatment of functional data as traditional multivariate data  can lead to poor performance since the algorithms are ignoring the correlation in the curvature of each function. In this paper we propose a method to analyse subspace structure of the functional data by using the state of the art Low-Rank Representation (LRR). Experimental evaluation on synthetic and real data reveals that this method massively outperforms conventional LRR in tasks concerning functional data.
\end{abstract}

\section{Introduction}

In machine learning it is common to interpret each data point as a vector in Euclidean space \cite{Bishop2006}. Such a discretisation is chosen because it allows for easy closed form solutions and fast computation, even with large datasets. However these methods ignore the fact that the data may not naturally fit into this assumption. In fact much of the data collected for practical machine learning are actually functions i.e. curves. For example financial data such as stock or commodity prices are functions of monetary value over time. Functional data have become increasingly important in many scientific and engineering research areas such as ECG (electrocardiogram) or EEG (Electroencephalography) in healthcare, biology data analysis, weather or climate data and motion trajectories from computer vision.  

Analyzing functional data has been an emerging topic in statistical research \cite{FerratyRomain2011,Mueller2011,SrivastavaShantanuJermyn2011,SrivastavaWuKurtekKlassenMarron2011}  and  has attracted great attention from machine learning community in recent years \cite{BahadoriKaleFanLiu2015,PetitjeanForestierWebbNicholsonChenKeogh2014}.   One of important challenges in analyzing functional data for machine learning is to efficiently cluster and to learn better representations for functional data. Theoretically the underlying process for functional data is of infinite dimension, thus it is difficult to work with them with only finite samples available.  A desired model for functional data is expected to properly and parsimoniously characterize the nature and variability hidden in the data. The classic functional principal component analysis (fPCA) \cite{RamsaySilverman2005} is one of such examples to discover dominant modes of variation in the data. However fPCA may fail to capture patterns if the functional data are not well aligned in its domain.  For time series, a special type of functional data, dynamic time warping (DTW) has long been proposed to compare time series based on shape and distortions (e.g., shifting and stretching) along the temporal axis \cite{Rakthanmanon2013,TuckerWuSrivastava2013}.

Another important type of functional data is shape \cite{SuSrivastavaHuffer2013,SrivastavaShantanuJermyn2011}.   Shape is an important characterizing feature for objects and in computer vision shape has been widely used for the purpose of object detection, tracking, classification, and recognition. In fact, a natural and popular representation for shape analysis is to parametrize boundaries of planar objects as 2D curves. In object recognition, images of the same object should be similar regardless of resolution, lighting, or orientation. Hence an efficient shape representation or shape analysis scheme must be invariant to scale, translation and rotation. A very useful shape representation is the square-root velocity function (SRVF) representation \cite{JoshiKlassenSrivastavaJermyn2007,SrivastavaShantanuJermyn2011}. In general, the resulting SRVF of a continuous shape is square integrable, a well-defined Hilbert space where appropriate measurement can be applied, refer to Section~\ref{Sec:2} for more details. By acknowledging the true nature of the data we can develop more powerful methods that exploit features that would otherwise be ignored or lead to erroneous results with simple linear models. 

%\cite{TuckerWuSrivastava2013}
%Unfortunately the reduction to vector based data points means that the expressive power of the models is rather limited. Take the example of subspace analysis which is the concern of this paper. Usually the models are linear in nature and each data point is considered as a linear combination of the other data points or of some other basis. This sort of global linear treatment of each data point discards invaluable information encoded in the curves. For example two curves have the same shape but one is the shift of another. This poses great challenge to these methods. 

Our intention in this study is to consider functional data clustering by accounting for the possible invariance in scaling/stretching, translation and rotation of functional data to help maintain shape characteristics. The focus of this paper is upon functional data where data sets consist of continuous real curves including shapes in Euclidean spaces. More specifically we propose a method of subspace analysis for functional data based on the idea developed in recent subspace clustering. The idea is to apply a feature mapping such as the aforementioned SRVF to the curves so that they are transformed onto the curve manifold, where the subspace analysis can be conducted based on the geometry on the manifold. In particular, we adapt the well known low-rank representation (LRR) framework \cite{LiuLinYanSunYuMa2013} to deal with data that lie on the manifold of open curves by implementing the classical LRR in tangent spaces of the manifold \cite{FuGaoHongTien2015,WangHuGaoSunYin2015,YinGaoGuo2015}. 

LRR on Euclidean spaces \cite{LiuLinYanSunYuMa2013} is closely related to several state-of-the-art subspace analysis approaches such as Sparse Subspace Clustering (SSC) \cite{ElhamifarVidal2013}, Robust PCA (RPCA) \cite{CandesLiMaWright2010} and low-rank Matrix Completion (MC) \cite{WuGaneshShiMatsushitaWangMa2012} methods.   %In pattern analysis it is necessary to define the underlying structure for the data at hand. Most commonly the model chosen works with linear subspaces since they afford easy computation and have shown practical success. It is well known that many types of data such as motion and face data can be well modelled by linear subspaces. For example Principal Component Analysis (PCA), Robust PCA and Matrix Completion methods \todo{Citations} all assume that the entire dataset is drawn from a single subspace. However in reality it is rare that a large dataset can be modelled accurately by a single subspace. Therefore it is much more sensible to considered that the data contains a mixture of several subspaces.\guoc{I found this logic is a bit odd. } \GaoC{I agree with Yi here. What we need to stress is the fact that some data are very special like functional data for which we are normally concerned with its shape and trends. It is this request we consider a model for this type of data. I suggest we shall first write the reasons for the necessity of analyzing this new type of data, then we talk about the possible SRV representation which actually paves the way by using the standard linear algorithms, then talk about the LRR.}
This mixture of subspaces model has naturally led to the development of subspace segmentation methods. Such methods aim to segment the data into clusters with each cluster corresponding to a unique subspace. More formally, given a data matrix of observed column-wise data samples $\mathbf A = [\mathbf{ a_1,a_2,\dots,a_N}] \in \mathbb{R}^{D \times N}$, the objective of subspace clustering is to assign each data sample to its underlying subspace. The basic assumption is that the data within $\mathbf A$ is drawn from a union of $c$ subspaces $\{S_i\}^c_{i=1}$ of dimensions $\{d_i\}^c_{i=1}$. 
%Both the number of subspaces $c$ and the dimension of each subspace may not be known.

The core of both SSC and LRR is to learn an affinity matrix for the given dataset and the learned affinity matrix will be pipelined to a spectral clustering method like nCUT \cite{ShiMalik2000} to obtain the final subspace labels. 
%In this work we limit discussion to spectral methods for subspace clustering. Spectral methods consist of two steps:
%\begin{enumerate}
%\item Learn a matrix which encodes the subspace structure;
%\item Apply spectral clustering to obtain subspace labels.
%\end{enumerate}
To learn the affinity matrix, SSC relies on the self expressive property \cite{ElhamifarVidal2013}, which is that\begin{quote}
{\it{each data point in a union of subspaces can be efficiently reconstructed by a linear combination of other points in the data}}.
\end{quote}
In other words, each point can be written as a linear combination of the other points i.e.\ $\mathbf{A = A Z}$, where $\mathbf Z \in \mathbb{R}^{N \times N}$ is a matrix of coefficients. Most methods however assume the data generation model $\mathbf{X = A + N}$, where $\mathbf X$ is the observed data and $\mathbf N$ is noise.
Since it is difficult to separate the noise from the data the solution is to relax the self-expressive model to  $\mathbf{X = X Z + E}$, where $\mathbf E$ is a fitting error and is different from $\mathbf N$.

Similarly LRR \cite{LiuLinYanSunYuMa2013} exploits the self expressive property but attempts to learn the global subspace structure by computing the lowest-rank representation of the set of data points. In other words, data points belonging to the same subspace should have similar coefficient patterns. In the presence of noise LRR attempts to minimise the following objective
\begin{align}
\min_{\mathbf{Z, E}} \; \frac{1}{2}\| \mathbf E \|_{\ell} + \textrm{rank}( \mathbf{Z} ), \quad
\text{s.t.} \quad \mathbf{X = XZ + E}.  \label{(1)}
\end{align}
However rank minimisation is an intractable problem. Therefore LRR actually uses the nuclear norm $\| \cdot \|_*$ (sum of the matrix's singular values) as the closest convex relation
\begin{align}
\min_{\mathbf{Z, E}} \; \frac{1}{2}\| \mathbf E \|_{\ell} + \| \mathbf{Z} \|_*, \quad
\text{s.t.} \quad \mathbf{X = XZ + E},  \label{(2)}
\end{align}
where $\| \cdot \|_{\ell}$ is a placeholder for the norm most appropriate to the expected noise type. For example in the case of Gaussian noise the best choice is the $\ell_2$ norm i.e.\ $\| \cdot \|_F^2$ and for sparse noise the $\ell_1$ norm should be used.

Both SSC and LRR rely on the linear self expressive property. This property is no longer available in the nonlinear manifold, e.g. the manifold of open curves as mentioned previously. To generalize LRR or SSC for data in the manifold space, we explicitly explore the underlying nonlinear data structure and utilize the techniques of exponential and logarithm mappings to bring data to a local linear space. 

The rest of the paper is organized as follows. In Section \ref{Sec:2}, we review the preliminaries about the manifold of open curves and introduce the curve Low-Rank Representation (cLRR) model. Section \ref{Sec:3} is dedicated to  explaining an efficient algorithm for solving the optimization proposed in cLRR based on the linearized alternative direction method with adaptive penalty (LADMAP) and the algorithm convergence and complexity are also analyzed.  In Section \ref{Sec:4}, the proposed model is assessed    on both synthetic and real world databases against several state-of-the-art methods. Finally, conclusions are discussed in Section \ref{Sec:5}.

\section{LRR over the Curve Manifold}\label{Sec:2}

As previously discussed LRR is limited to a linear model and its current version can only be applied to vector data from a Euclidean space. Matrix $\mathbf Z$ in \eqref{(1)} or \eqref{(2)} encodes the affinity/similarity between data points. However this assumption is often unnatural and quite limiting. Much of the data encountered in real world is functional. In other words it exhibits a curve like structure over a domain. Euclidean linear models are unable to capture the nonlinear invariance embedded in each data point. For example in thermal infra-red data of geological substances a curve may contain a key identifying feature such as a dip near a particular frequency. This dip may shift or vary position over time even for the same substance due to impurities. Under a linear vector model this variation may cause the vector to drastically move in the ambient Euclidean space and cause poor results. Or in other cases the feature may be elongated, shrunk or be subject to some non-uniformly warping or scaling. In all these cases the linear model will fail to accurately represent the non-linear affinity in the data.

Exploring these unique non-linear invariance in functional data is the focus of this paper. We now discuss how to adapt LRR (similar approach appliable to SSC) such that it easily accepts curve data and nonlinear relationships within clusters can be easily discovered.
\subsection{The Curve Manifold}
Given a smooth parameterized $n$-dimension curve $\beta : D = [0, 1] \to \mathbb R^n$, we represent it using he square-root velocity function (SRVF) representation  \cite{JoshiKlassenSrivastavaJermyn2007,SrivastavaShantanuJermyn2011}, which is given by
\begin{align*}
q(t) = \frac{\dot{\beta}(t)}{\sqrt{\| \dot{\beta}(t) \|}}.
\end{align*}
The SRVF mapping transforms the original curve $\beta(t)$ into a gradient based representation, which facilitates the comparing of the shape information. 

In this paper, we focus on the set of open curves, e.g. the curves do not form a loop ($\beta(0) \neq \beta(1)$). For handling general curves, we refer readers to  \cite{SrivastavaShantanuJermyn2011}. The SRVF facilitates a measure and geometry bearing invariance to scaling, shifting and reparameterization in the curves domain. For example, all the translated curves from a curve $\beta(t)$ will have the same SRVF.  Robinson \cite{Robinson2012} proved that if the curve $\beta(t)$ is absolutely continuous, then its SRVF $q(t)$ is square-integrable, i.e., $q(t)$ is in a functional Hilbert space $L^2(D, \mathbb{R}^n)$ .  Conversely for each $q(t)\in L^2(D, \mathbb{R}^n)$, there exists a curve $\beta(t)$ whose SRVF corresponds to $q(t)$. Thus the set $L^2(D, \mathbb{R}^n)$ is a well-defined representation space of all the curves.  The most important advantage offered by the SRVF framework is that the natural and widely used $L^2$-measure on $L^2(D, \mathbb{R}^n)$ is invariant to the reparameterization. That is, for any two SRVFs $q_1$ and $q_2$ and a randomly chosen reparametrization function (non-decreasing) $t = \gamma(\tau)$, we have 
\[
\|q_1(t) - q_2(t)\|_{L^2} = \|q_1(\gamma(\tau)) - q_2(\gamma(\tau))\|_{L^2}.
\]

This property has been exploited in \cite{BahadoriKaleFanLiu2015} for functional data clustering under the subspace clustering framework. Different from the work proposed in \cite{BahadoriKaleFanLiu2015}, we will adopt the newly developed LRR on manifolds framework to the model of curves LRR, see \cite{FuGaoHongTien2015,WangHuGaoSunYin2015,YinGaoGuo2015}. To see this, we introduce some more notation.  Let $\Gamma$ be the set of all diffeomorphisms from $D=[0,1]$ to $D=[0,1]$. This set collects all the reparametrization mappings. $\Gamma$ is a Lie group with the composition as the group operation and the identity mapping as the identity element. Then all the orbits $[q] = \{ q\circ \gamma = q(\gamma(t)) \;|\; \forall \gamma\in \Gamma\}$ together define the quotient manifold $L^2(D, \mathbb{R}^n)/\Gamma$. %, on which we will extend the standard LRR to.  

Without loss of generality, all curves are normalized to have unit length, i.e., $\int_{D}\|\dot{\beta}(t)\|dt = 1$. The SRVFs associated with these curves are elements of a unit hypersphere in the Hilbert space $L^2(D, \mathbb{R}^n)$, i.e., $\int_D \| q(t) \|^2 dt = 1$.  Therefore, under the curve normalization assumption, instead of $L^2(D, \mathbb{R}^n)$, we consider the following unit  hypersphere manifold
%Given these assumptions and the SRV of each curve,  we can say that each curve $q$ is an element of the manifold of open curves $\mathcal C^o$\todo{Citation}. It is known that $\mathcal C^o$ forms a unit hypersphere in the Hilbert Manifold $L^2(D, \mathbb R^n)$
\begin{align*}
\mathcal C^o = \bigg\{ q \in L^2(D, \mathbb R^n): \int_D \| q(t) \|^2 dt = 1 \bigg\}.
\end{align*}

The manifold $\mathcal C^o$ has some nice properties, see \cite{AbsilMahonySepulchre2008}.  For any two points $q_0$ and $q_1$ in $\mathcal C^o$, a
geodesic connecting them is given by $\alpha: [0, 1] \rightarrow \mathcal C^o$,
\begin{align}
\alpha (\tau)  = \frac1{\sin(\theta)} (\sin(\theta(1 -\tau))q_0 + \sin(\theta\tau)q_1), 
\end{align}
where $\theta = \cos^{-1}(\langle q_0 , q_1 \rangle)$ is the length of the geodesic. If we take derivative of $\alpha$ w.r.t to $q_1$, the tangent vector
at $q_0$ is
\begin{align}
v = \frac{\theta}{\sin(\theta)}[q_1 - \langle q_0 , q_1 \rangle q_0]. \label{Tangent1}
\end{align} 
The above formula is regarded as the {\t Logarithm} mapping $\log_{q_0} (q_1)$ on the manifold $\mathcal C^o$.

As we are concerned with the shape invariance, i.e., we need to additionally remove the shape-preserving transformations: rotation and curve reparametrization. The manifold concerning us is the quotient space of the manifold $\mathcal S^o = \mathcal C^o/(SO(n) \times \Gamma)$, where $SO(n)$ is the rotation group. Each element $[q]\in \mathcal{S}^o$ is an equivalent class defined by 
\[
[q] = \left\{O q(\gamma(t))\sqrt{\dot{\gamma}(t)}\; |\; O\in SO(n) \text{ and } \gamma \in\Gamma \right\}.
\]

Given any two points $[q_0]$ and $[q_1]$ in $\mathcal S^o$,  a  tangent representative \cite{AbsilMahonySepulchre2008} in the tangent space $T_{[q_0]} (\mathcal S^o)$ can be calculated in the following way, as suggested in \cite{ZhangSuKlassenLeSrivastava2015,SuSrivastava2014} based on \eqref{Tangent1},
\begin{align}
\widetilde{v} = \log_{q_0} (\widetilde{q}_1) = \frac{\widetilde{\theta}}{\sin(\widetilde{\theta})}[\widetilde{q}_1 - \langle q_0 , \widetilde{q}_1 \rangle q_0]. \label{Tangent2}
\end{align}
where $\widetilde{q}_1$ is the representative of $[q_1]$ given by the well-defined algorithm in \cite{ZhangSuKlassenLeSrivastava2015,SuSrivastava2014} and $\widetilde{\theta} = \cos^{-1}(\langle q_0 , \widetilde{q}_1 \rangle)$. In fact, $\widetilde{v}$ is the lifting representation of abstract tangent vector $\log_{[q_0]}([q_1])$ on $T_{[q_0]}(\mathcal{S}^o)$ at $q_1$.

\subsection{The Proposed Curve LRR}
Given a set of $N$ unit-length curves $\{\beta_1(t), ..., \beta_N(t)\}$, denote their SRVFs by $\{q_1(t), ..., q_N(t)\}$ such that $[q_i]\in \mathcal S^o$ and $q_i(t)$ is a representative of the equivalent class $[q_i]$. We cannot apply the standard LRR model \eqref{(2)} directly on the quotient manifold $\mathcal S^o$. This is because \eqref{(2)} indeed relies on the following individual linear combination
\begin{align}
\mathbf x_i = \sum^N_{j=1} z_{ij} \mathbf x_j + \mathbf e_i, \label{linear}
\end{align}
which is invalid for $[q_i]$'s on $\mathcal S^o$.   Note that $z_{ij}$ can be explained as the affinity or similarity between data points $\mathbf{x}_i$ and $\mathbf{x}_j$.

On any manifold, the tangent space at a given point is linearly local approximation to the manifold around the point and the linear combination is valid in the tangent space. This prompts us to replace the affinity relation in \eqref{linear} by the following 
\begin{align}
\log_{[q_i]}([q_i]) = \sum_{j=1}^N w_{ij} \log_{[q_i]} ([q_j]) + \mathbf{e}_i\label{tangentLinear}
\end{align}
with the constraint $\sum_{j=1}^N w_{ij} = 1, i = 1, 2, \dots, N$ to maintain consistency at different locations. The meaning of $w_{ij}$ in \eqref{tangentLinear} is the similarity between curves $\beta_i(t)$ and $\beta_j(t)$ via the ``affinity'' between tangent vectors $\log_{[q_i]}([q_i]) $ and $\log_{[q_i]}([q_j])$ at the first order approximation accuracy.  Each $\log_{[q_i]}([q_j])$ can be calculated by \eqref{Tangent2} and it is obvious that $\log_{[q_i]}([q_i])=0$ for any $i$.

%Since our data lies on a hypersphere (i.e. a manifold in Hilbert functional space) rather than in a Euclidean space we can no longer use the linear self representation of LRR. However we can form an approximation of this manifold in around a data point by its tangent space using a log mapping. The log mapping $\log_{\mathbf q_i}(\mathbf q_j)$ maps curve $j$ in the local tangent space of curve $i$. Thus the self representation model becomes
%
%Note that
%\begin{align*}
%\log_{\mathbf q_i}(\mathbf q_i) = \mathbf 0
%\end{align*}
%since $\mathbf q_i$ is the origin vector of the tangent space defined by log mapping $\log_{\mathbf q_i}$.
%
%Fortunately the log mapping is easily computed \todo{citation} as
%\begin{align*}
%\log_{\mathbf{q}_i}(\mathbf{q}_j) = \frac{\theta}{\sin(\theta)} [\mathbf{q}_j - \langle \mathbf{q}_i, \mathbf{q}_j \rangle \mathbf{q}_i ]
%\end{align*}
%where $\theta = \cos^{-1}( \langle \mathbf{q}_i, \mathbf{q}_j \rangle)$ is the length of the geodesic distance between $\mathbf{q}_i$ and $\mathbf{q}_j$). However to ensure that an accurate log mapping is obtained we first subject each $\mathbf{q}_i$ to re-parameterisation to align the curves. For the sake of brevity we refer readers to \cite{srivastava2011shape} for full details of the alignment process.

With all the ingredients at hand, we are fully equipped to propose the curve LRR (cLRR) model as follows 
\begin{align}
\begin{aligned}
\min_{\mathbf W} \lambda \| \mathbf W \|_* + \sum_{i = 1}^N \frac{1}{2} \| \sum_{j=1}^N w_{ij} \log_{[q_i]} ([q_j]) \|_{[q_i]}^2, \\
\textrm{s.t.} \; \sum_{j=1}^N w_{ij} = 1, i = 1, 2, \dots, N.
\end{aligned}\label{Model}
\end{align}
where $\|\cdot\|_{[q]}$ is the metric defined on the manifold, which is defined by the classic $L^2$ Hilbert metric on the tangent space.  

Denote $\mathbf w_i$ the $i$-th row of matrix $\mathbf W$ and define
\begin{align} 
\label{b_create}
B^i_{jk} = \langle \log_{[q_i]} ([q_j]), \log_{[q_i]} ([q_k]) \rangle.
\end{align}
Then with some algebraic manipulation we can re-write the model \eqref{Model} into the following simplified form, 
\begin{align}
\begin{aligned}
\min_{\mathbf W} \lambda \| \mathbf W \|_* + \sum_{i = 1}^N \mathbf w_i \mathbf B^i \mathbf w_i^T, \\
\textrm{s.t.} \; \sum_{j=1}^N w_{ij} = 1, i = 1, 2, \dots, N.
\end{aligned}\label{obj_curve}
\end{align}
where $\mathbf B^i  = (B^i_{jk})$.
 
Effectively this objective allows for similarity between curves to be measured in their tangent spaces. Our highly accurate segmentation results in Section \ref{Sec:4} have demonstrated that this is an effective way to learn  non-linear similarity.

\section{Optimisation}\label{Sec:3}
\subsection{Algorithm}
To solve the cLRR objective we use the Linearized Alternative Direction Method with Adaptive Penalty (LADMAP) \cite{LinLiuLi2015,LinLiuSu2011} . First take the Augmented Lagrangian of the objective \eqref{obj_curve}
\begin{equation}
\begin{aligned}
L = &\lambda \|\mathbf W\|_*  + \frac12\sum^N_{i=1}\mathbf w_i \mathbf{B}^i\mathbf w^T_i + \langle {\mathbf y}, \mathbf W\mathbf 1 - \mathbf 1\rangle \\
&+ \frac{\beta}2\|\mathbf W\mathbf 1 - \mathbf 1\|^2_F
\end{aligned}\label{25August2014-5}
\end{equation}
where $\mathbf y$ is the Lagrangian multiplier (vector) corresponding to the equality constraint $\mathbf W\mathbf 1 = \mathbf 1$, $\|\cdot\|_{F}$ is the matrix Frobebius-norm, and we will update $\beta$ as well in the iterative algorithm to be introduced.

Denote by $F(\mathbf W)$ the function defined by \eqref{25August2014-5} except for the first term $\lambda \|\mathbf W\|_*$. To solve \eqref{25August2014-5}, we adopt a linearization of $F(\mathbf{W})$ at the current location $\mathbf W^{(k)}$ in the iteration process, that is, we approximate $F(\mathbf W)$ by the following linearization with a proximal term
\begin{align*}
F(\mathbf W)\approx & F(\mathbf W^{(k)}) + \langle \partial F(\mathbf W^{(k)}), \mathbf W-\mathbf W^{(k)}\rangle \\
&+ \frac{\eta_{W}\beta_k}2\|\mathbf W-\mathbf W^{(k)}\|^2_F,
\end{align*}
where $\eta_W$ is an approximate constant with a suggested value given by $\eta_W = \max\{\|B_i\|^2\}+N+1$, and $\partial F(\mathbf W^{(k)})$ is a gradient matrix of $F(\mathbf W)$ at $\mathbf W^{(k)}$. Denote by $\mathbf B$ the 3-order tensor whose $i$-th front slice is given by $\mathbf B^i$. Let us define $\mathbf W\odot \mathbf B$ the matrix whose $i$-row is given by $\mathbf w_i \mathbf B^i$, then it is easy to show
\begin{equation} 
\partial F(\mathbf W^{(k)}) = \mathbf W\odot \mathbf B + \mathbf y\mathbf 1^T + \beta_k (\mathbf W\mathbf 1 - \mathbf 1)\mathbf 1^T.
\label{Eq:14October2014-4}
\end{equation}

%and $N$ is actually equal to the squared operator norm of $\mathcal{A}: %W\in\mathbb{R}^{N\times N} \rightarrow \mathcal{R}^N$ mapping the sum of each row of $W$ to 1.
Then \eqref{25August2014-5} can be approximated by linearization and $\mathbf w$ will be updated by the following
\begin{align}
 \mathbf W^{(k+1)}   %= &\arg\min _W F(W^{(k)}) + \langle \partial F(W^{(k)}), W-W^{(k)}\rangle +\frac{\eta_B\beta_k}{2}\|W-W^{(k)}\|^2_F + \lambda \|W\|_*\\
= &\arg\min_{\mathbf W} \lambda \|\mathbf W\|_* \label{SolutionW}\\
+&\frac{\eta_W\beta_k}{2} \bigg\|\mathbf W - \left(\mathbf W^{(k)} - \frac{1}{\eta_W\beta_k} \partial F(\mathbf W^{(k)})\right)\bigg\|^2_F. \notag
\end{align}

\begin{algorithm}[]
\caption{{\bf Solving \eqref{obj_curve} by LADMAP}}
\label{curve_lrr_alg}
\begin{algorithmic}[1]

\REQUIRE $\{\mathbf X_i\}_{i=1}^N$, $\lambda$

\STATE Initialise: $\mathbf W = \mathbf 0$, $\mathbf y = \mathbf 0$, $\beta = 0.1$, $\beta_{\text{max}} = 10$, $\rho^0 = 1.1$, $\eta = \max \{ \| \mathbf B^i \|_F \} + N + 1$, $\epsilon_1 = 1e^{-4}$, $\epsilon_2 = 1e^{-4}$

\STATE Construct each $\mathbf B^i$ as per \eqref{b_create}

\WHILE{not converged}

\STATE Update $\mathbf W$ using \eqref{SolutionWk}

\STATE Check convergence criteria
\begin{align*}
\beta^{(k)} \| \mathbf W^{(k+1)} - \mathbf W^{(k)} \|_F \leq \epsilon_1\\
\| \mathbf W \mathbf 1 - \mathbf 1 \|_F \leq \epsilon_2 \
\end{align*}

\STATE Update Lagrangian Multiplier
\begin{align*}
\mathbf y^{(k+1)} = \mathbf y^k + \beta^{(k)} ( \mathbf W \mathbf 1 - \mathbf 1 )^T
\end{align*}

\STATE Update $\rho$
\begin{align*}
\rho = 
\begin{cases}
\rho_0 & \text{if} \;\; \beta^{(k)} \| \mathbf W^{(k+1)} - \mathbf W^{(k)} \|_F \leq \epsilon_1 \\
1 & \text{otherwise,}
\end{cases}
\end{align*}

\STATE Update $\beta$
\begin{align*}
\beta^{(k+1)} = \textrm{min}(\beta_{\textrm{max}}, \rho \beta^{(k)})
\end{align*}
 
\ENDWHILE

\RETURN $\mathbf W$

\end{algorithmic}
\end{algorithm}

\begin{figure*}[!th]
\centering
	\subfloat[Cluster 1]{
	\includegraphics[width=0.3\linewidth]{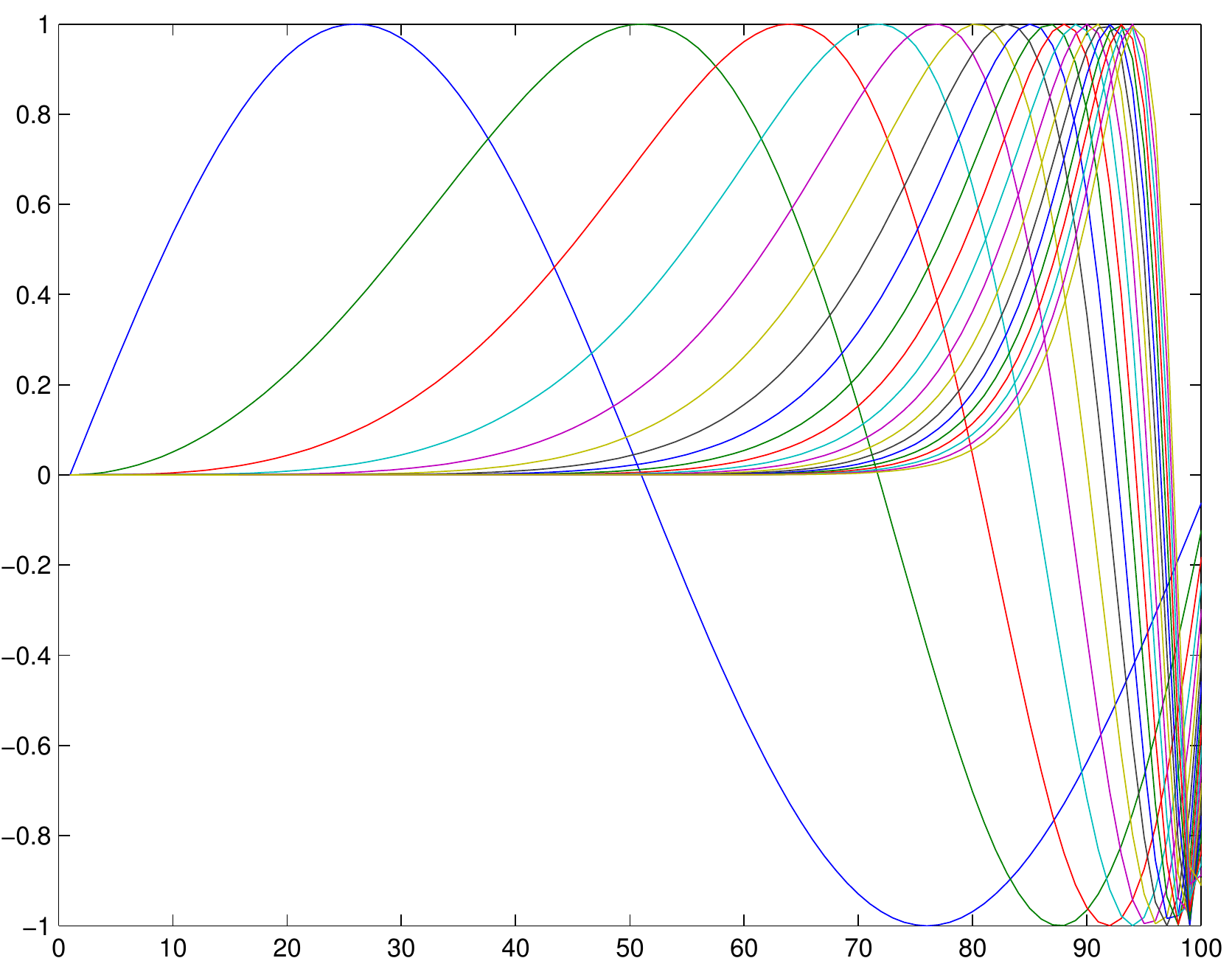}}
	\subfloat[Cluster 2]{
	\includegraphics[width=0.3\linewidth]{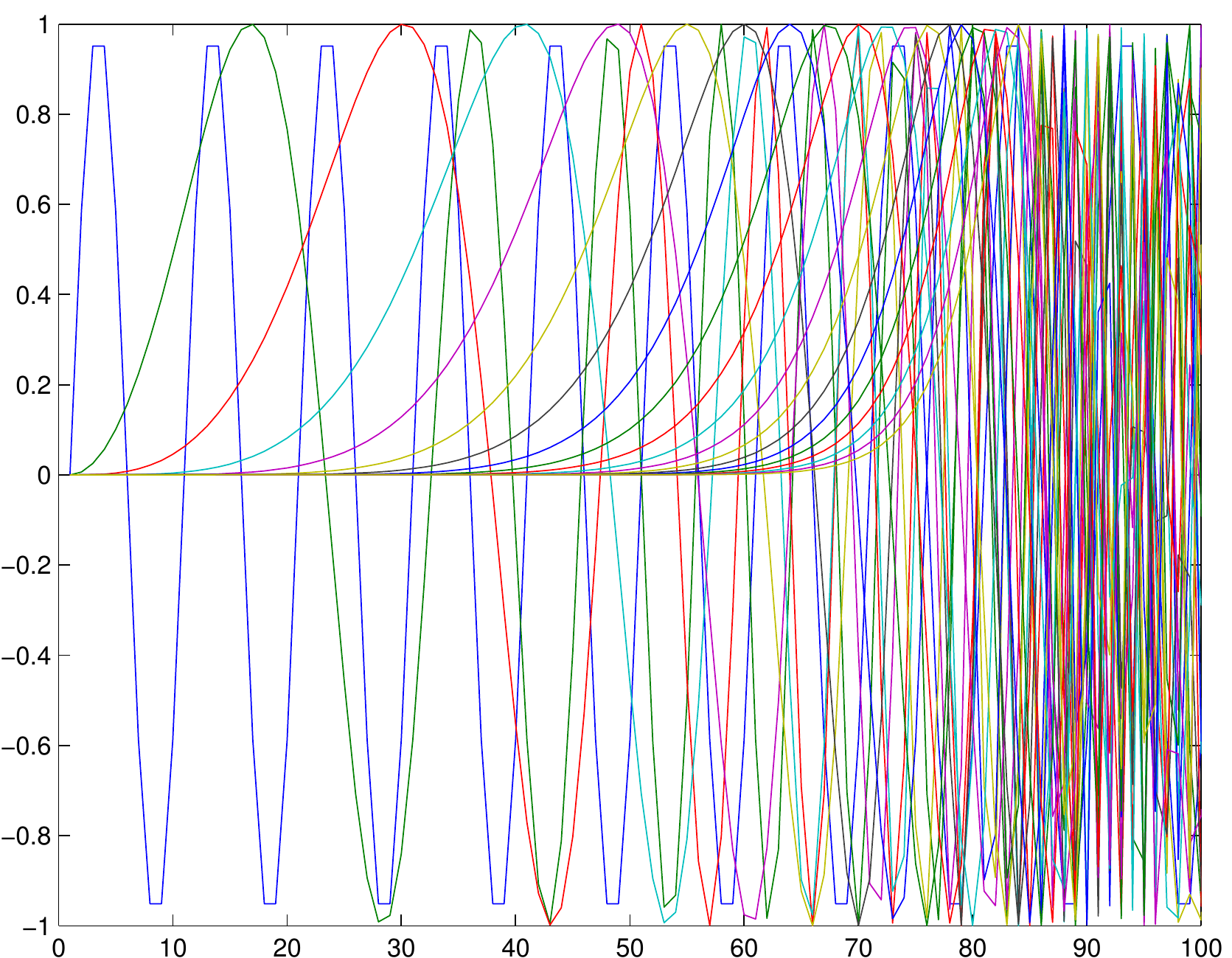}}
	\subfloat[Cluster 3]{
	\includegraphics[width=0.3\linewidth]{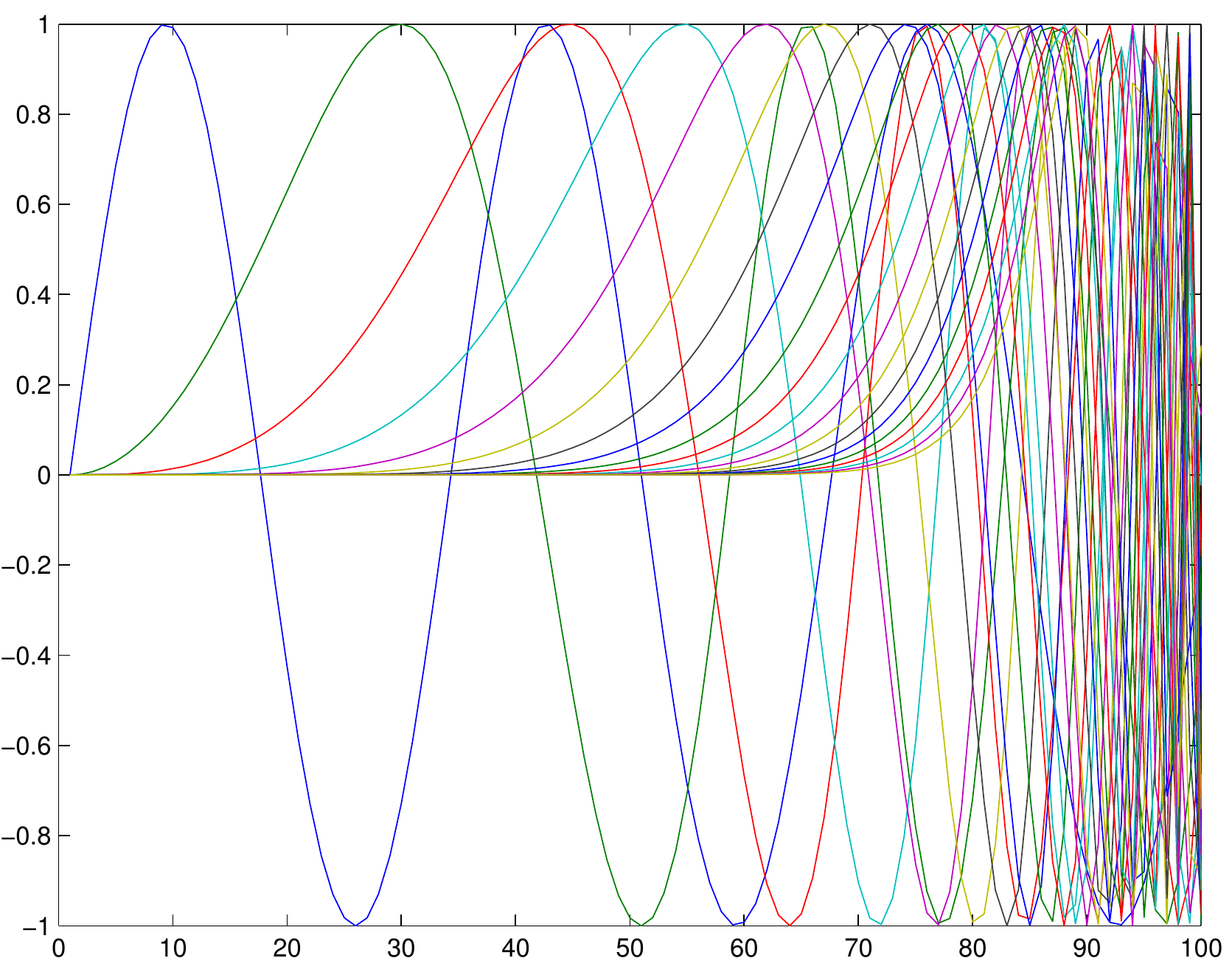}}
	
\caption{Example plots of curves generated in the Synthetic Data Experiment. Each cluster has a base sine curve (the left most blue curve) which is progressively warped with each successive instantiation.}
\label{fig_syn_data}
\end{figure*}

Problem \eqref{SolutionW} admits a closed form solution by using SVD thresholding operator \cite{CaiCandesShen2008}, given by
\begin{equation}\label{SolutionWk}
\begin{aligned}
\mathbf W^{(k+1)} = U_{W} S_{\frac{\lambda}{\eta_W\beta_k}}(\Sigma_{W})V_{W}^{T},
\end{aligned}
\end{equation}
where $U_{W}\Sigma_{W}V_{W}^{T}$ is the SVD of $\mathbf W^{(k)} - \frac{1}{\eta_W\beta_k} \partial F(\mathbf W^{(k)})$ and $S_\tau(\cdot)$ is the Singular Value Thresholding (SVT) \cite{CaiCandesShen2008,parikh2013proximal} operator defined by
\begin{equation}
\begin{aligned}
S_{\tau}(\Sigma) = \text{diag}(\max\{|\Sigma_{ii}|-\tau, 0\}).
\end{aligned}
\end{equation}

The updating rule for $\mathbf y$
\begin{equation} 
\mathbf y^{(k+1)} = \mathbf y^{(k)} + \beta_k (\mathbf W^{(k)}\mathbf 1 -\mathbf 1) 
\label{Eq:14October2014-5}
\end{equation}
and the updating rule for $\beta_k$
\begin{equation}
\begin{aligned}
\beta_{k+1} = \min\{\beta_{\text{max}}, \rho \beta_k\},
\end{aligned}\label{UpdateBeta}
\end{equation}
where
 \[
 \rho = \begin{cases} \rho_0 & \beta_k \|\mathbf W^{k+1} - \mathbf W^k\| \leq \varepsilon_1,\\
 1 & \text{otherwise}.
 \end{cases}
 \]

We summarize the above as Algorithm~\ref{curve_lrr_alg}.  Once the coefficient matrix $\mathbf W$ is found, a spectral clustering like nCUT \cite{ShiMalik2000} is applied on the affinity matrix $\frac{|\mathbf W|+|\mathbf W|^T}{2}$ to obtain the segmentation of the data.

\subsection{Complexity Analysis}
For ease of analysis, we firstly define some symbols used in the following. Let $K$ and $r$ denote the total number of iterations and the lowest rank of the matrix $\mathbf W$, respectively. The size of $\mathbf W$ is $N\times N$. The major computation cost of our proposed method contains two parts, calculating all $\mathbf B^i$'s and updating $\mathbf W$. In terms of the formula \eqref{b_create} through \eqref{Tangent1} and \eqref{Tangent2}, the computational complexity of Log algorithm is $O(T^2)$ where $T$ is the number of terms in a discretized curves; therefore, the complexity of $B_{jk}^i$ is at most $O(T^2)$ and $\mathbf B^i$'s computational complexity is $O(N^2T^2)$. Thus the total for all the $\mathbf B^i$ is $O(N^3)$. In each iteration of the Algorithm, the singular value thresholding is adopted to update the low rank matrix $\mathbf W$ whose complexity is $O(rN^2)$~\cite{LiuLinYanSunYuMa2013}. Suppose the algorithm is terminated after $K$ iterations, the overall computational complexity is given by
\[
O(N^3)+O(KrN^2)
\]

\subsection{Convergence Analysis}
Algorithm~\ref{curve_lrr_alg} is adopted from the algorithm proposed in~\cite{LinLiuSu2011}. However due to the terms of $\mathbf B^i$'s in the objective function~\eqref{25August2014-5}, the convergence theorem proved in~\cite{LinLiuSu2011} cannot be directly applied to this case as the linearization is implemented on both the augmented Lagrangian terms and the term involving $\mathbf B^i$'s. Fortunately we can employ the revised approach, presented in \cite{YinGaoLinShiGuo2015}, to prove the convergence for the algorithm. Without repeating all the details, we present the convergence theorem for Algorithm~\ref{curve_lrr_alg} as follows.
\begin{theorem}[Convergence of Algorithm~\ref{curve_lrr_alg}] If $\eta_W\geq \max\{\|B_i\|^2\}+N+1$, $\displaystyle\sum^{+\infty}_{k=1}\beta^{-1}_k = +\infty$, $\displaystyle\beta_{k+1}-\beta_k > C_0 \frac{\sum_i \|B_i\|^2}{\eta_W - \max\{\|B_i\|^2\}-N}$, where $C_0$ is a given constant and $\|\cdot\|$ is the matrix spectral norm, then the sequence $\{W^{k}\}$ generated by Algorithm~\ref{curve_lrr_alg} converges to an optimal solution to problem~\eqref{obj_curve}.
\end{theorem}

In all the experiments we have conducted, the algorithm converges very fast with $K<100$.

\section{Experiments}\label{Sec:4}
In this section we show three sets of experiments to evaluate the newly proposed cLRR. The performance of the proposed method is compared with the same type of subspace clustering algorithm LRR \cite{LiuLinYanSunYuMa2013}. To compare segmentation accuracy we use the subspace clustering accuracy (SCA) metric \cite{ElhamifarVidal2013}, which is defined as
\begin{align}
\text{SCA} = 1 - \frac{\text{num. of misclassified points}}{\text{total num. of points}}.
\end{align}
Therefore a higher SCA $\%$ means greater clustering accuracy.

The parameters used were fixed across all experiments with $\lambda$ for LRR set at $1$ and $0.1$ for cLRR. A wide range of parameters were tested for each algorithm. Overall we found that the segmentation accuracy of LRR did not vary that much with changes in $\lambda$.

\subsection{Synthetic Data}

\begin{figure}[!h]
\centering
	\subfloat[LRR]{
	\includegraphics[width=1\linewidth]{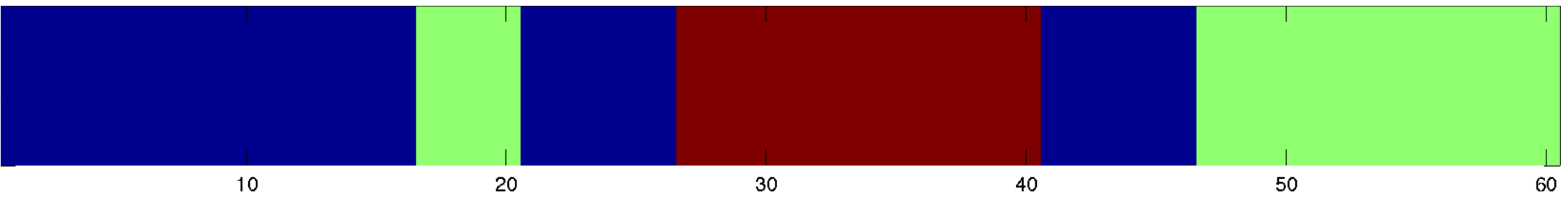}}\\
	\subfloat[Curve LRR]{
	\includegraphics[width=1\linewidth]{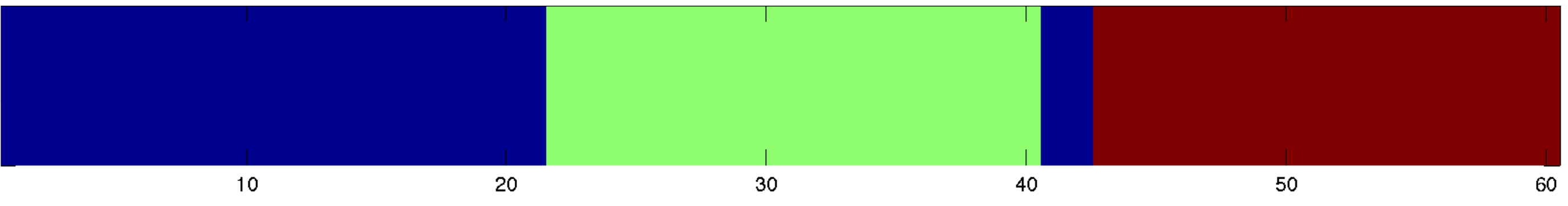}}
	
\caption{The segmentation results from the data in Figure \ref{fig_syn_data}.}
\label{fig_syn_clusters}
\end{figure}

\begin{table}[!h]
\centering

\begin{tabular}{c c c c c}
\hline
 & Mean & Median & Min & Max \\
\hline

LRR			& 80.4\%		& 83.33\%		& 60\%		& 91.67\% \\
CurveLRR	& \bf 96.77\%		& \bf 98.33\%		& \bf 73.33\%		& \bf 100\%
	
\end{tabular}

\caption{Synthetic Results}
\label{table_syn_results}
\end{table}

\begin{figure*}[htb]
\centering
	\subfloat[Cluster 1]{
	\includegraphics[width=0.24\linewidth]{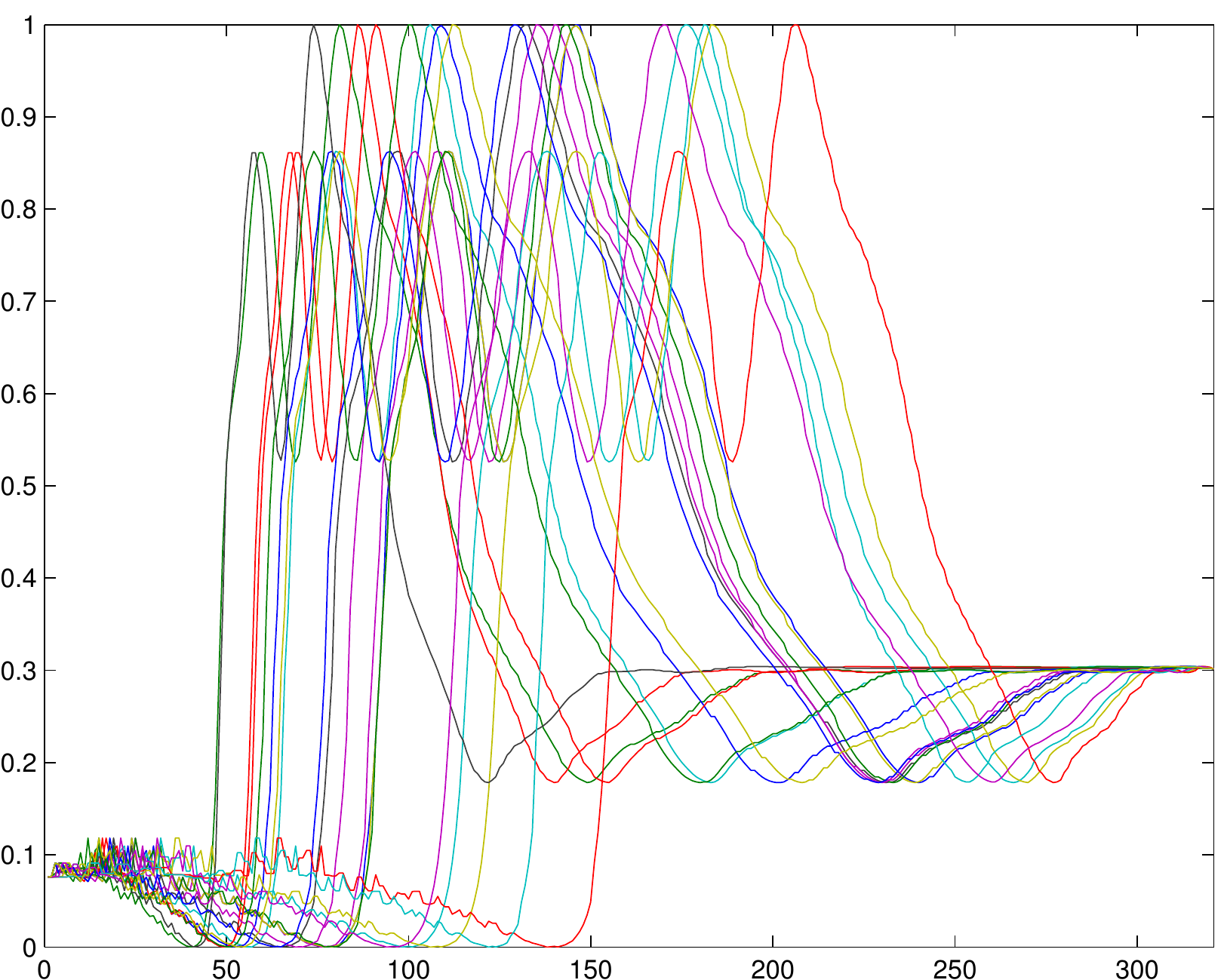}}
	\subfloat[Cluster 2]{
	\includegraphics[width=0.24\linewidth]{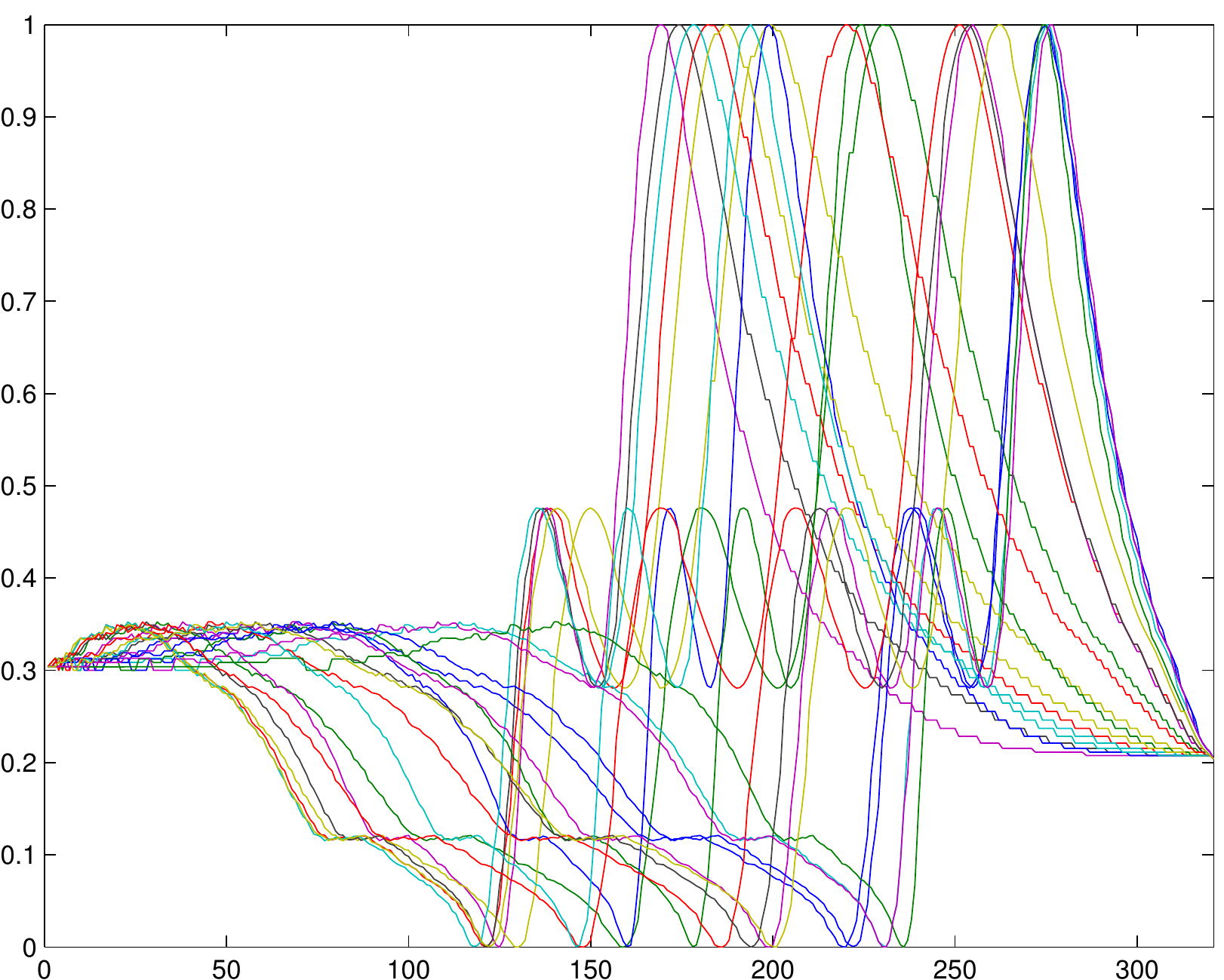}}
	\subfloat[Cluster 3]{
	\includegraphics[width=0.24\linewidth]{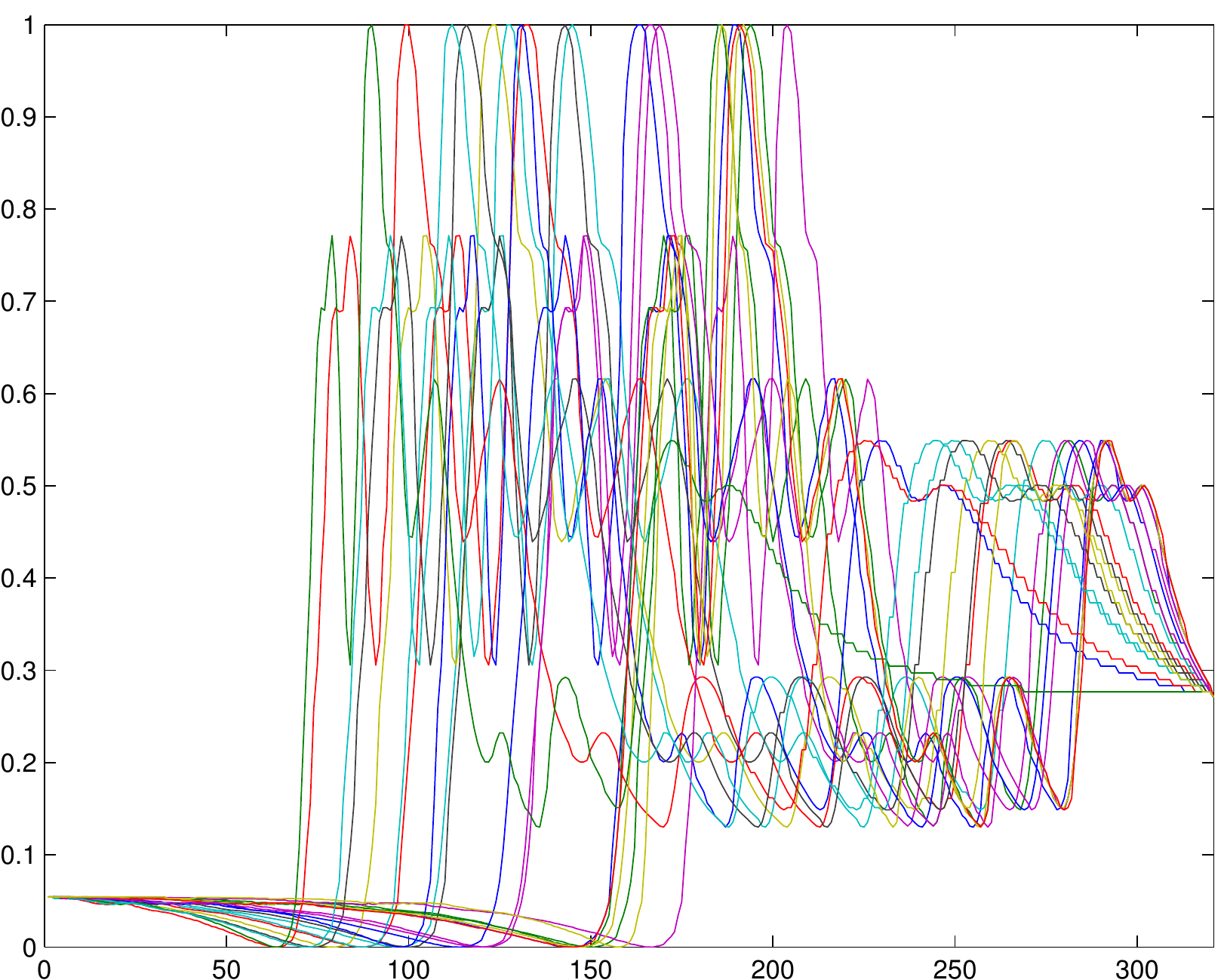}}
	\subfloat[Base Curves]{
	\includegraphics[width=0.24\linewidth]{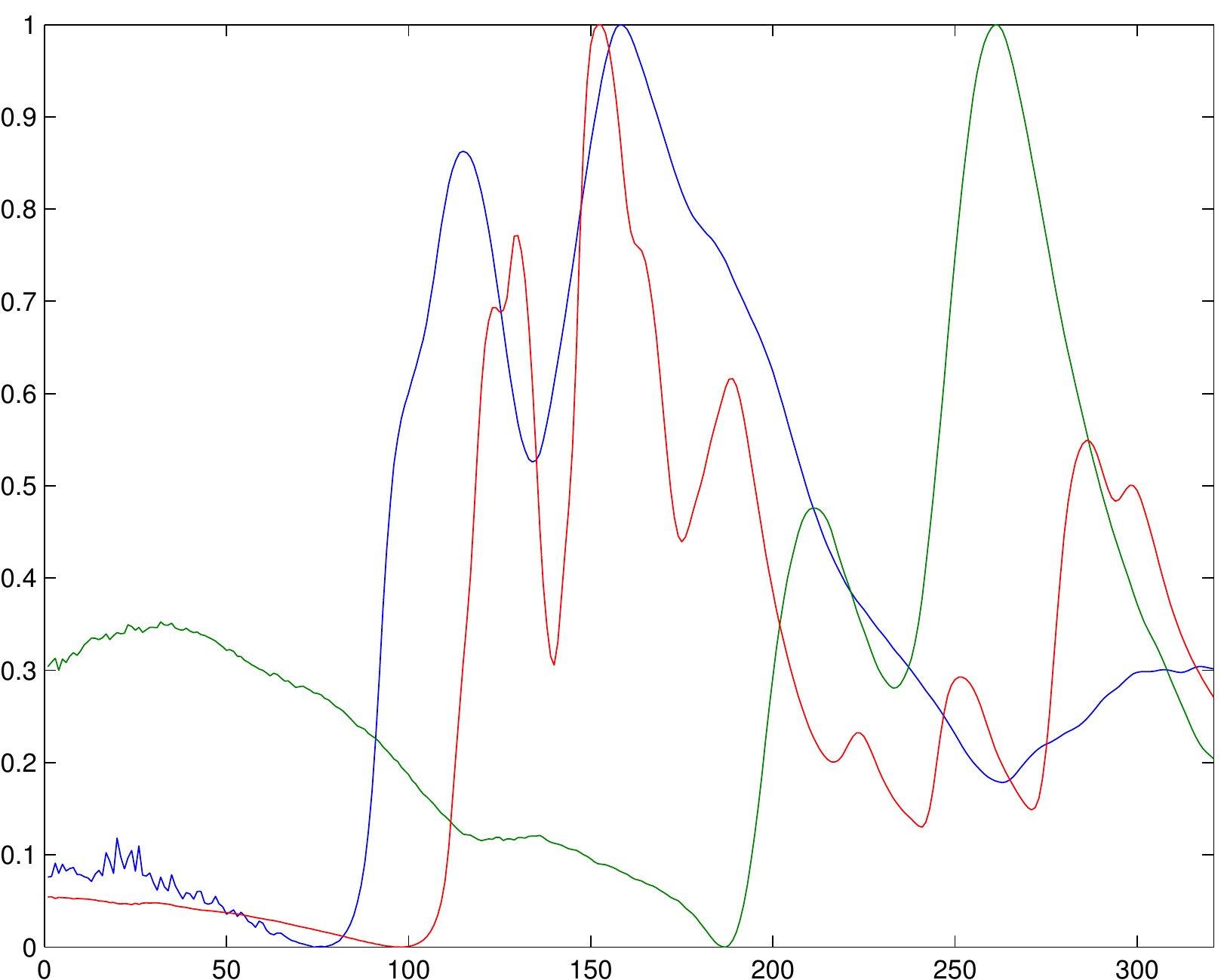}}
	
\caption{Example plots of curves used in the Semi-synthetic TIR Data Experiment. Each cluster has a base curve from the TIR library. The curves for each cluster have been shifted and stretched randomly from the base.}
\label{fig_tir_data}
\end{figure*}

To evaluate and confirm the effectiveness of the curve LRR method we first perform experimental evaluation using synthetic data. In this test three clusters were created consisting of twenty 1-D curves of length 100. The curves in each cluster were sine waves, with each cluster corresponding to a unique frequency. Within each cluster progressive amounts of warping were applied. See Figure \ref{fig_syn_data} for an example of data from three syntheticly generated clusters. Clustering was then performed on the data by applying curve LRR and segmenting the affinity matrix with nCUT. This experiment was repeated $50$ times with new data generated each time to obtain basic statistics. We compare against the baseline: LRR. Results are reported using subspace clustering accuracy and can be found in Table \ref{table_syn_results}. Overall in this experiment Curve LRR outperforms conventional LRR by a significant margin.

\subsection{Semi-synthetic TIR Data}

\begin{figure}[H]
\centering
	\subfloat[LRR]{
	\includegraphics[width=1\linewidth]{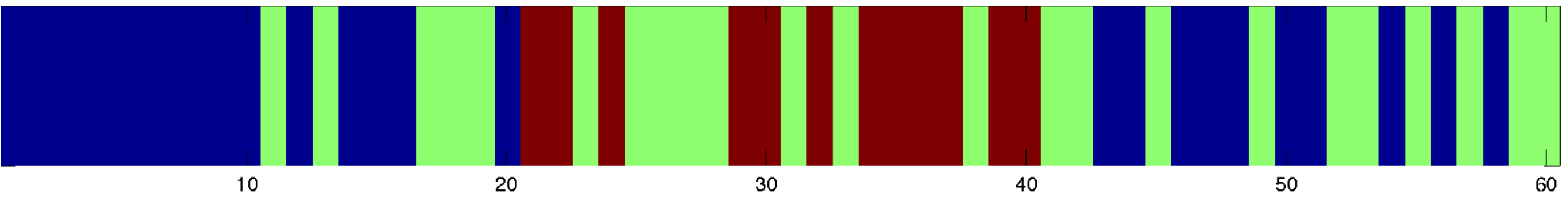}}\\
	\subfloat[Curve LRR]{
	\includegraphics[width=1\linewidth]{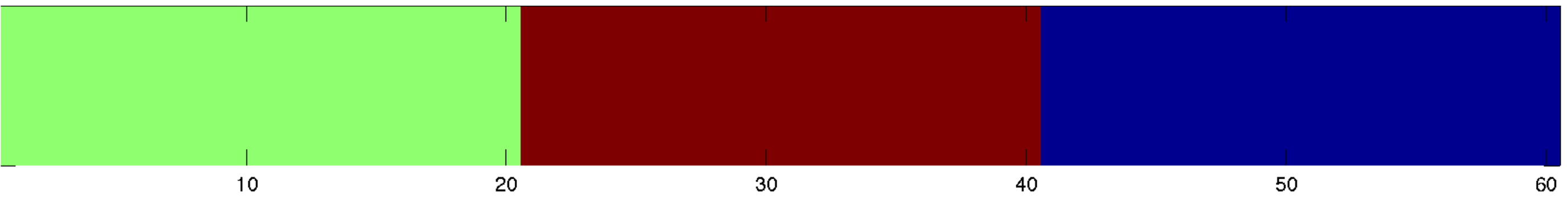}}
	\caption{The segmentation results from the data in Figure \ref{fig_tir_data}.}
\label{fig_tir_clusters}
\end{figure}

We assemble synthetic data from a library of pure infrared hyper spectral mineral data. For each cluster we pick one spectra sample from the library as a basis. Each curve basis is then randomly shifted and stretched in a random portion. This random warping is performed $20$ times to produce the curves for each cluster. See Figure \ref{fig_tir_data} for an example of data used in this experiment. In this experiment we used three clusters. Again as in the previous experiment we repeated the test $50$ times. Results are reported in Table \ref{table_tir_results} and Figure \ref{fig_tir_clusters}.

The results show that LRR cannot accurately cluster data with this sort of nonlinear invariance, which is commonly found in this type of data due to impurities in the mineral samples. On the other hand cLRR perfectly clustered the data.

\begin{table}[!h]
\centering

\begin{tabular}{c c c c c}
\hline
 & Mean & Median & Min & Max \\
\hline

LRR			& 60.13\%		& 60\%		& 50\%		& 71.67\% \\
CurveLRR	& \bf 100\%		& \bf 100\%		& \bf 100\%		& \bf 100\%
	
\end{tabular}

\caption{Semi-Synthetic TIR Results}
\label{table_tir_results}
\end{table}

\subsection{Character Classification}

\begin{figure*} 
\centering
	\subfloat[Trajectories for ``a'']{
	\includegraphics[width=0.25\linewidth]{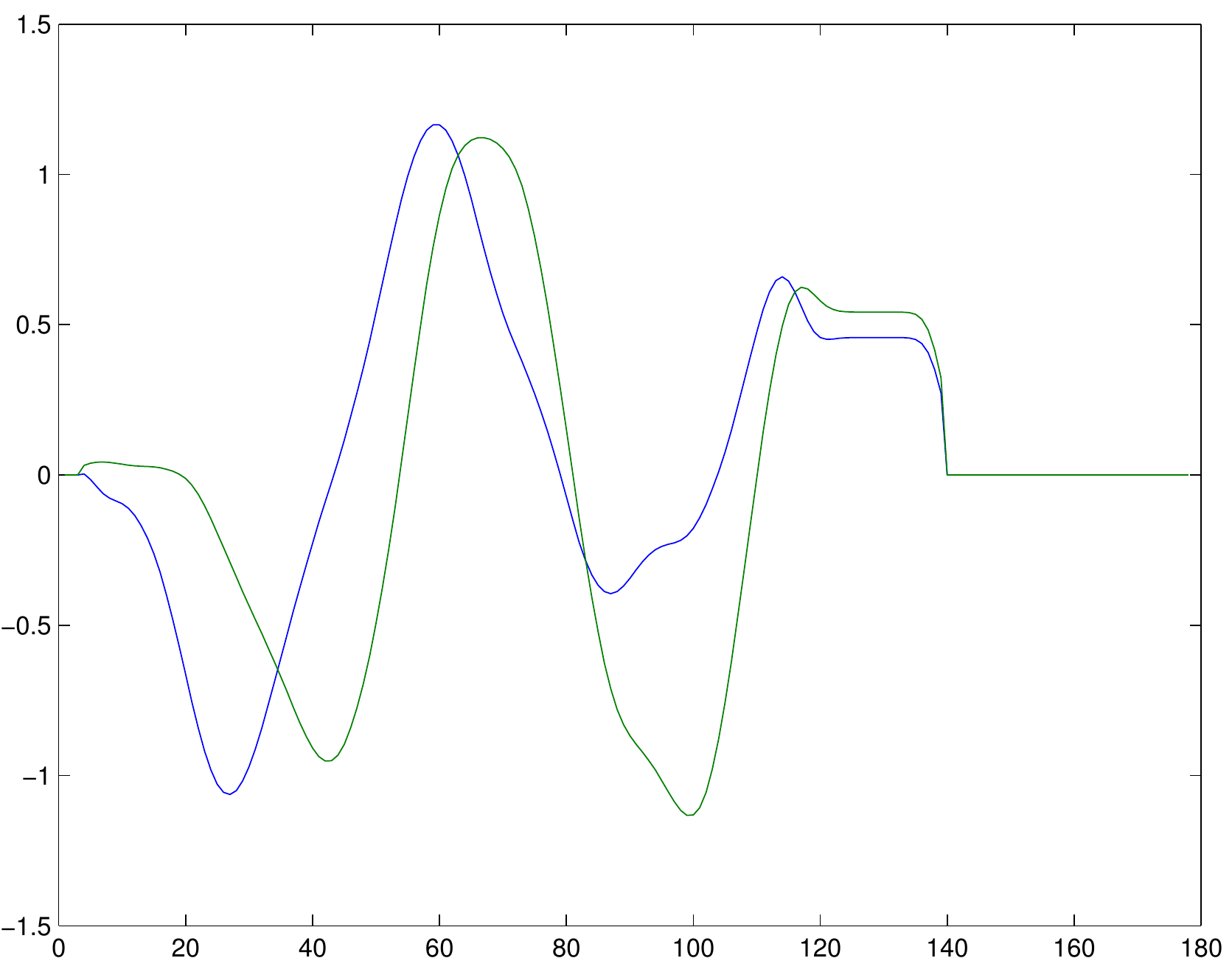}}
	\subfloat[Trajectories for ``b'']{
	\includegraphics[width=0.25\linewidth]{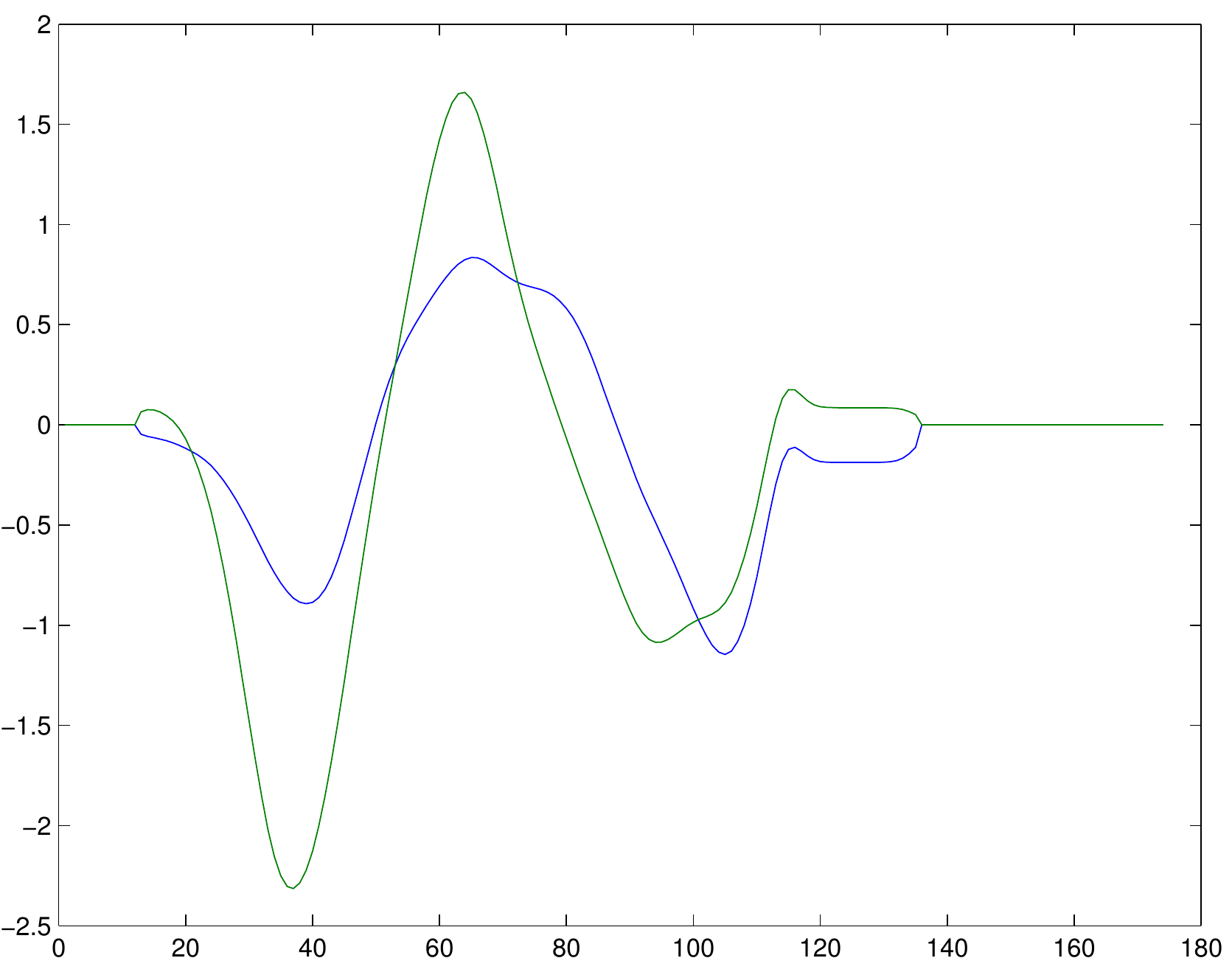}}
	\subfloat[Trajectories for ``c'']{
	\includegraphics[width=0.25\linewidth]{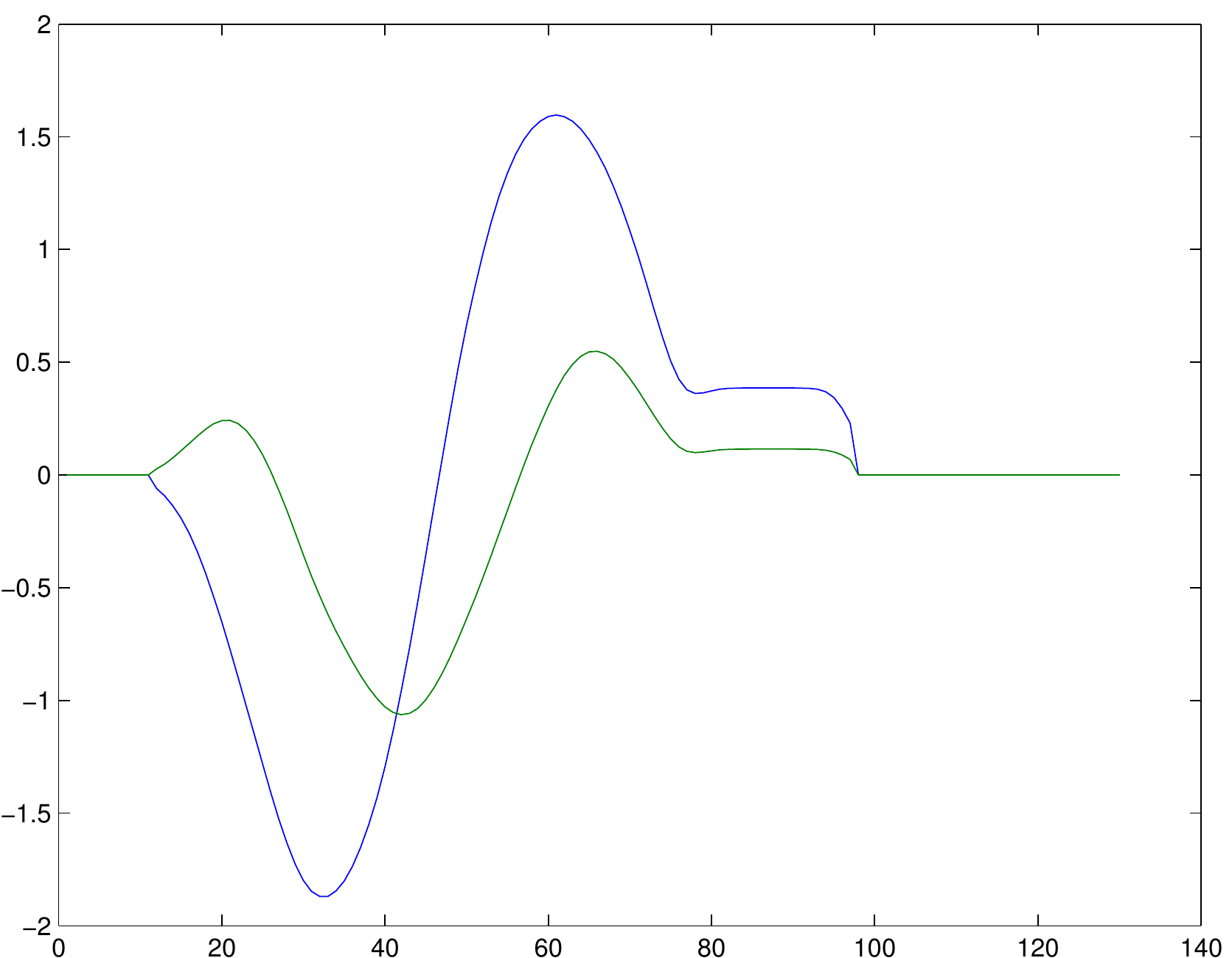}}\\
	\subfloat[Reconstructed ``a'']{
	\includegraphics[width=0.25\linewidth]{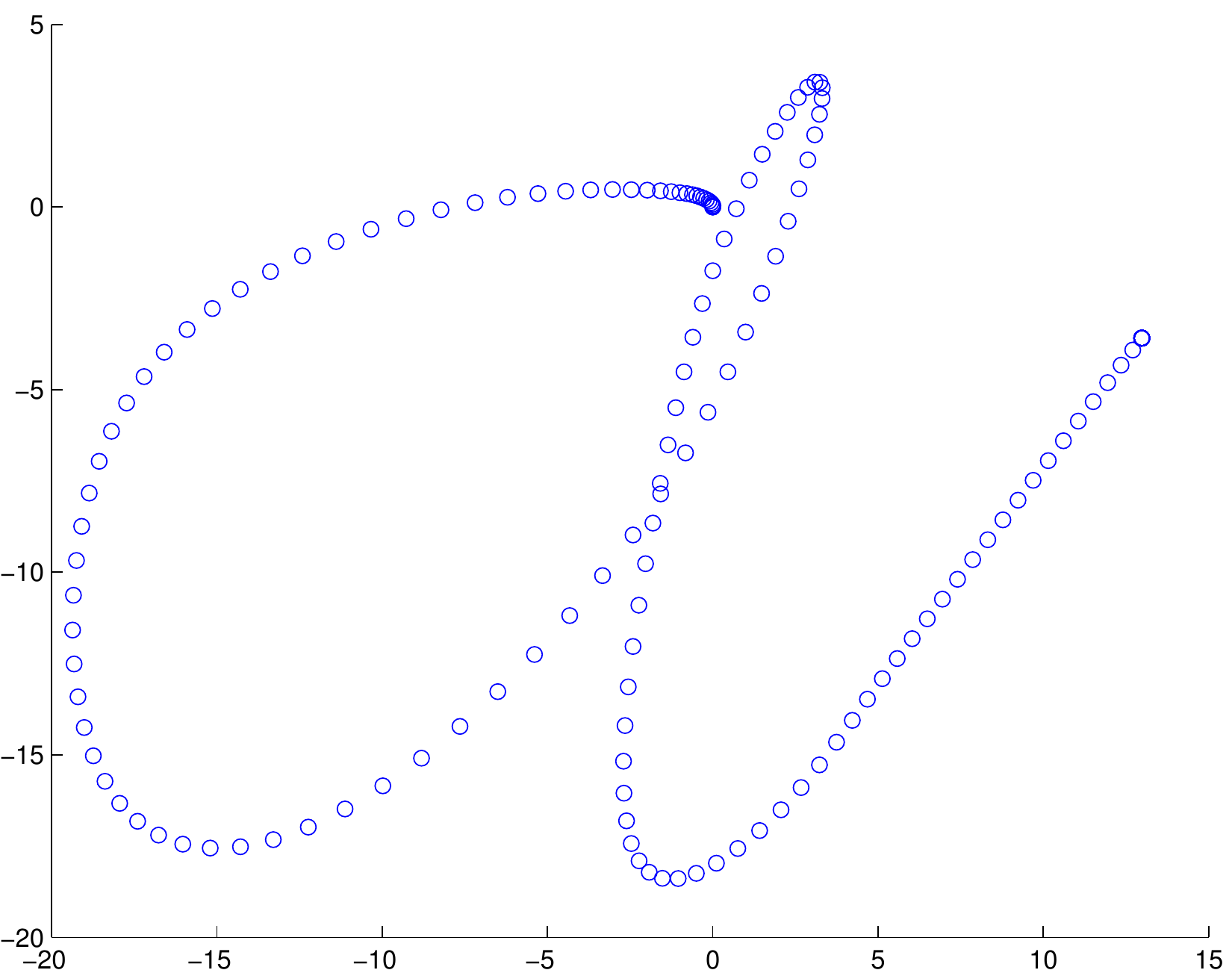}}
	\subfloat[Reconstructed ``b'']{
	\includegraphics[width=0.25\linewidth]{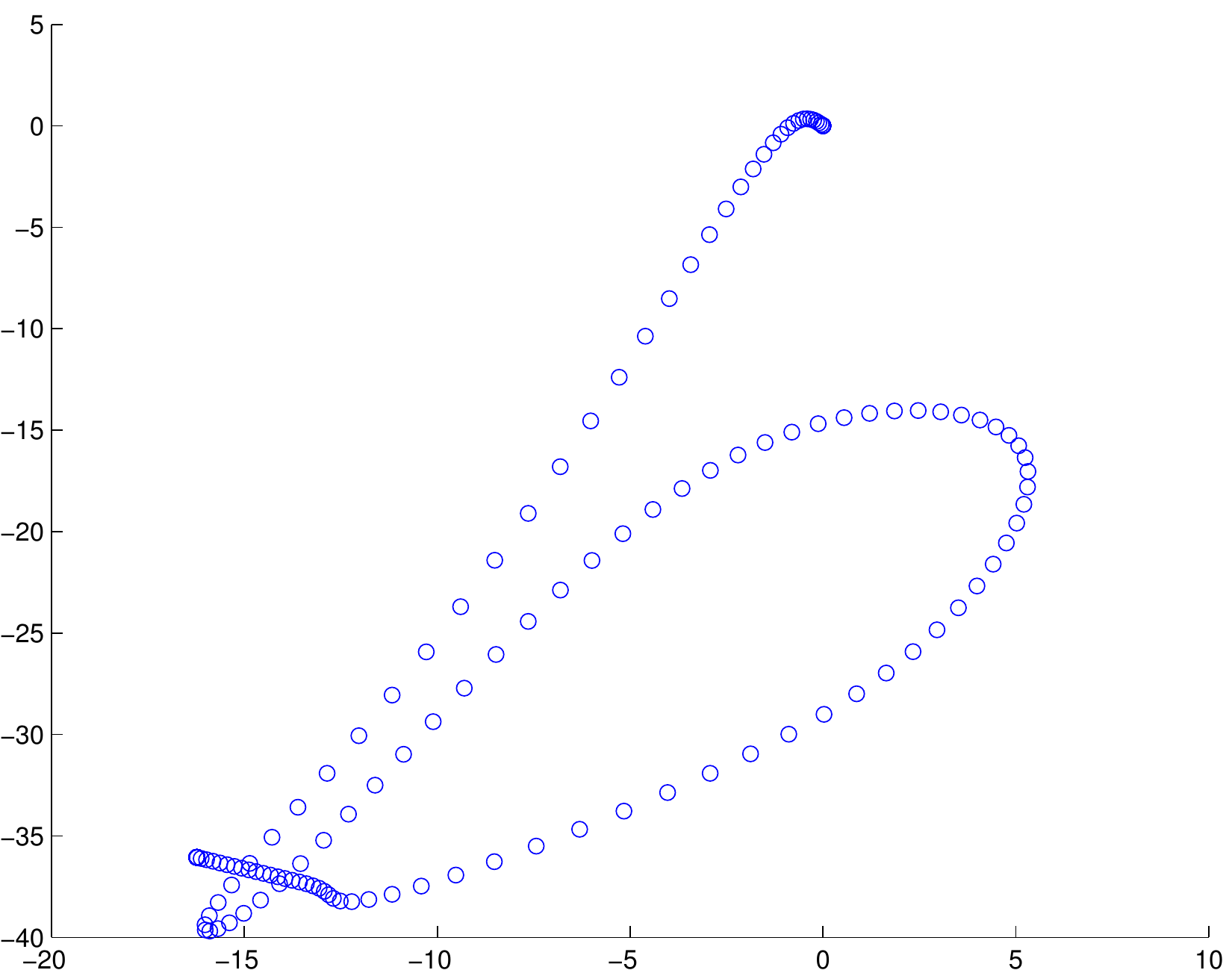}}
	\subfloat[Reconstructed ``c'']{
	\includegraphics[width=0.25\linewidth]{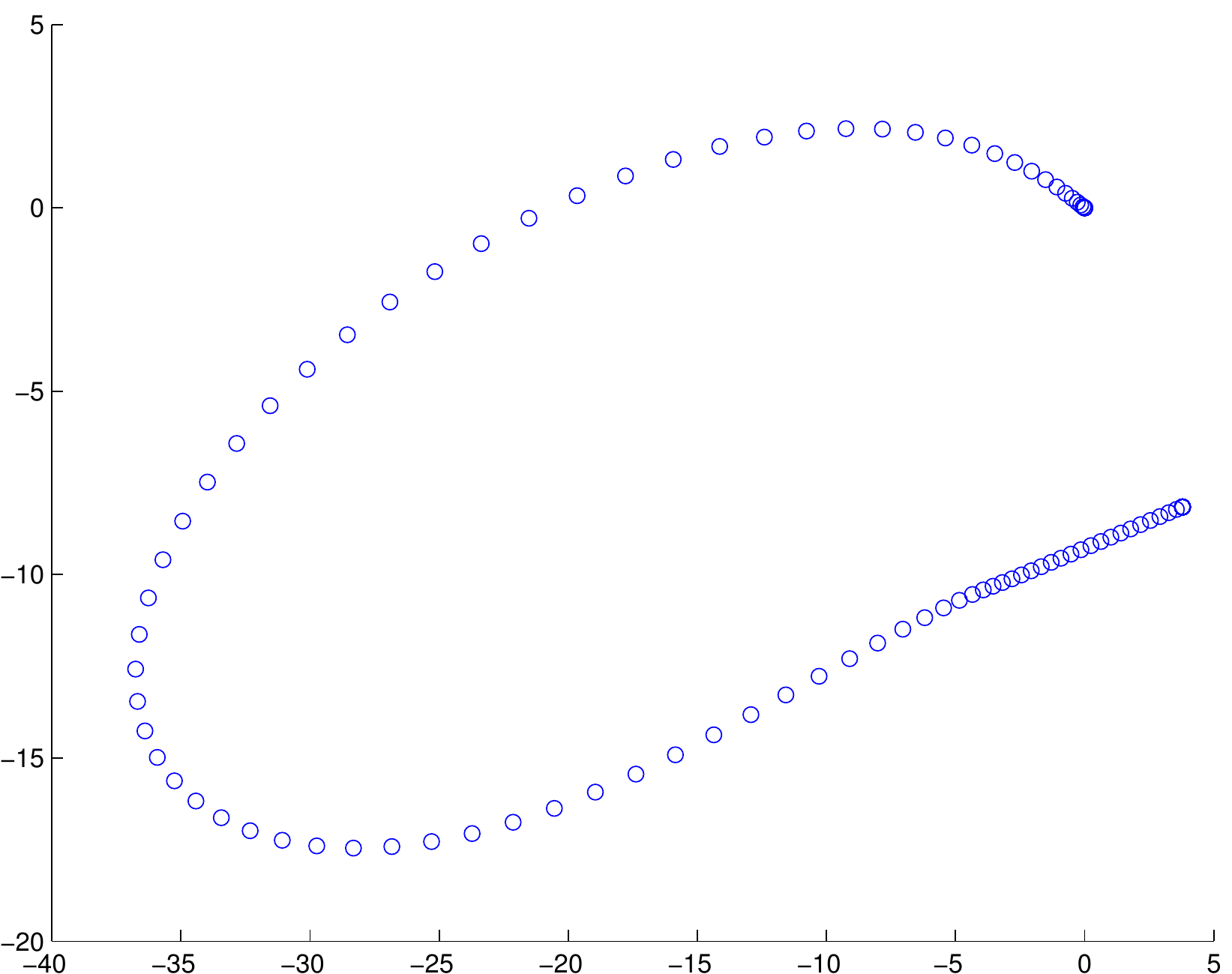}}
\caption{Example data from the character classification dataset. The top row plots the x and y pen tip velocities over time for three sample characters. The bottom row shows the corresponding character reconstruction by integrating the pen tip velocity data (for visualisation only).}
\label{fig_char_examples}
\end{figure*}

\begin{figure*}
\centering
	\subfloat[Cluster 1 - X]{
	\includegraphics[width=0.27\linewidth]{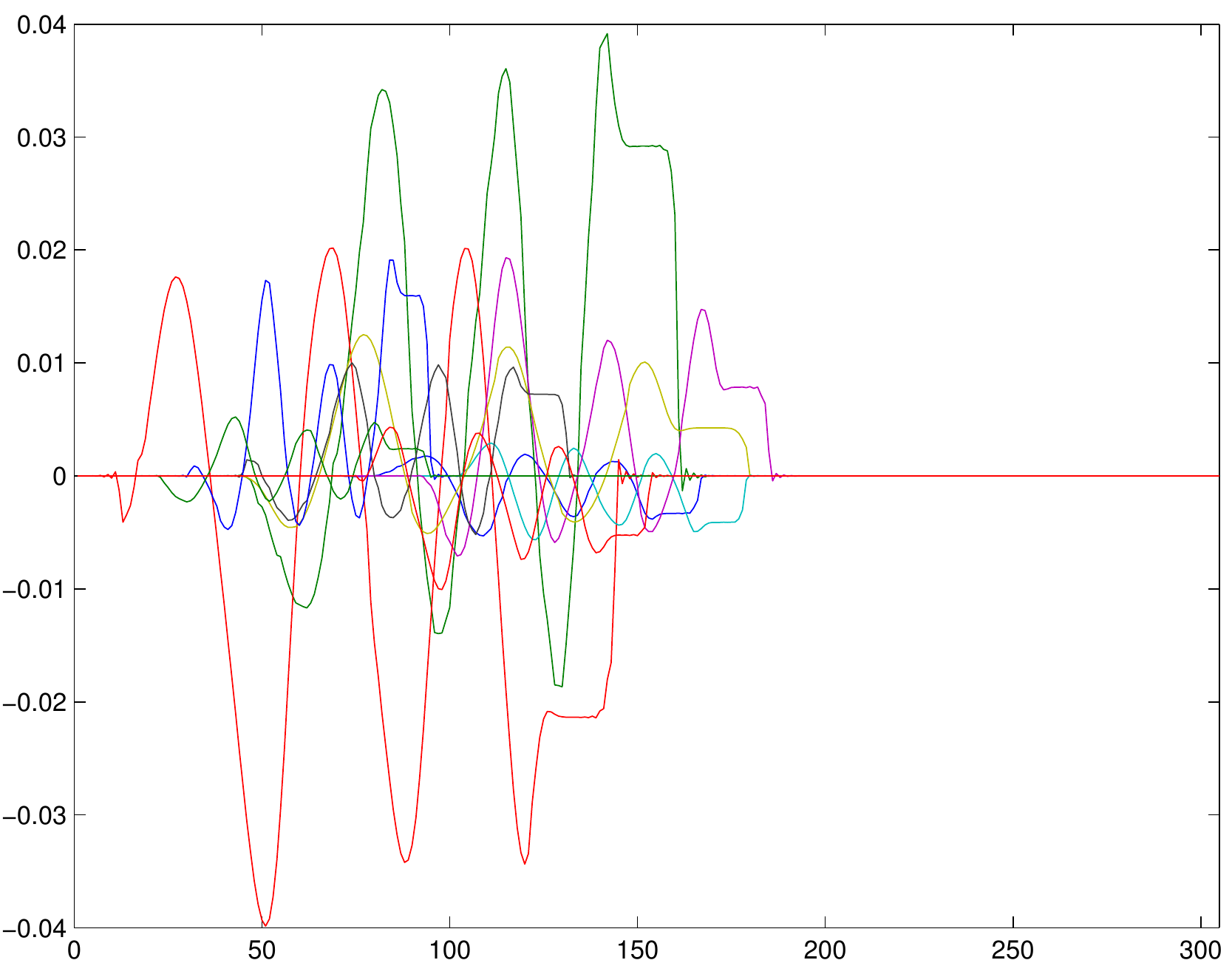}}
	\subfloat[Cluster 2 - X]{
	\includegraphics[width=0.27\linewidth]{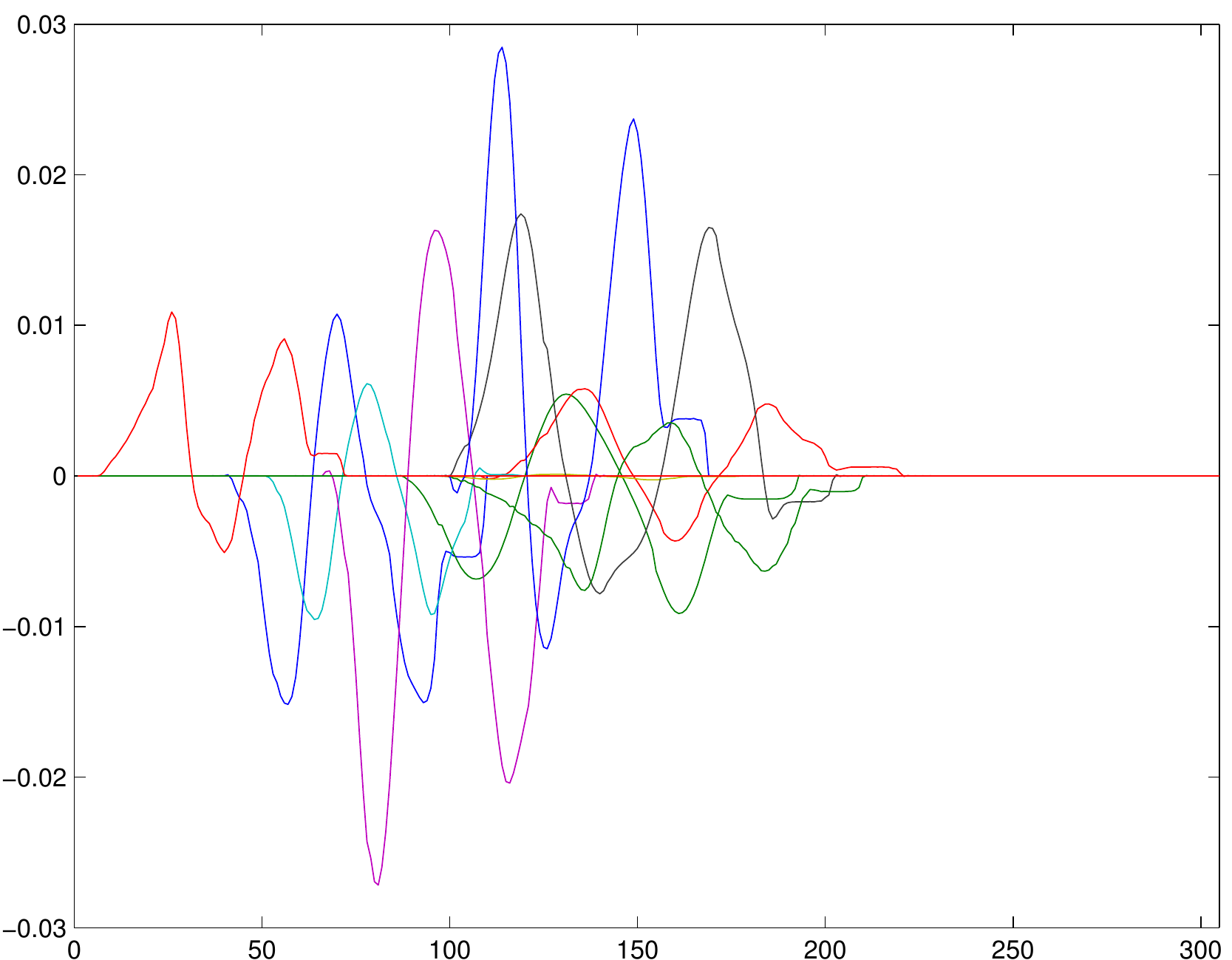}}
	\subfloat[Cluster 3 - X]{
	\includegraphics[width=0.27\linewidth]{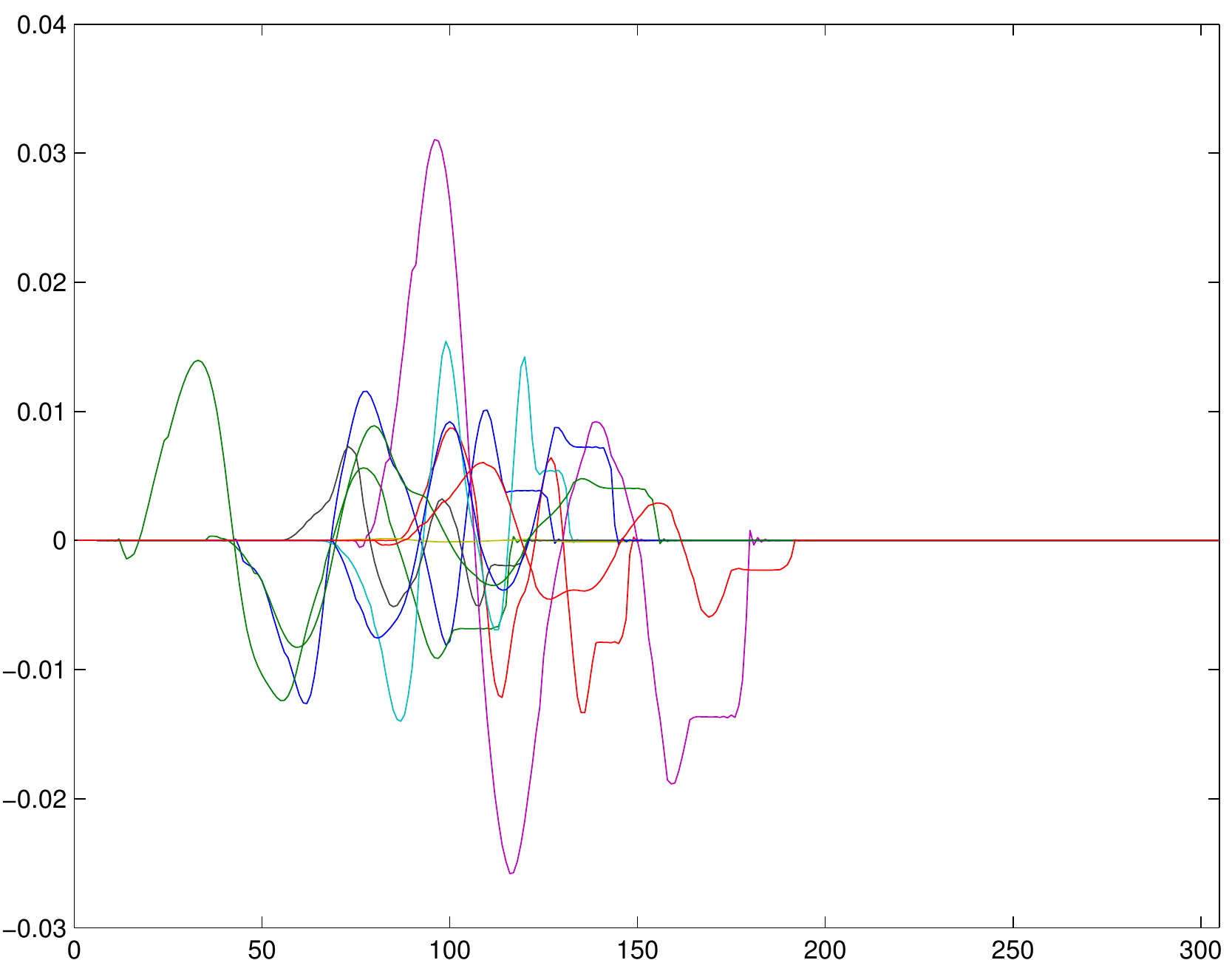}}\\
	\subfloat[Cluster 1 - Y]{
	\includegraphics[width=0.27\linewidth]{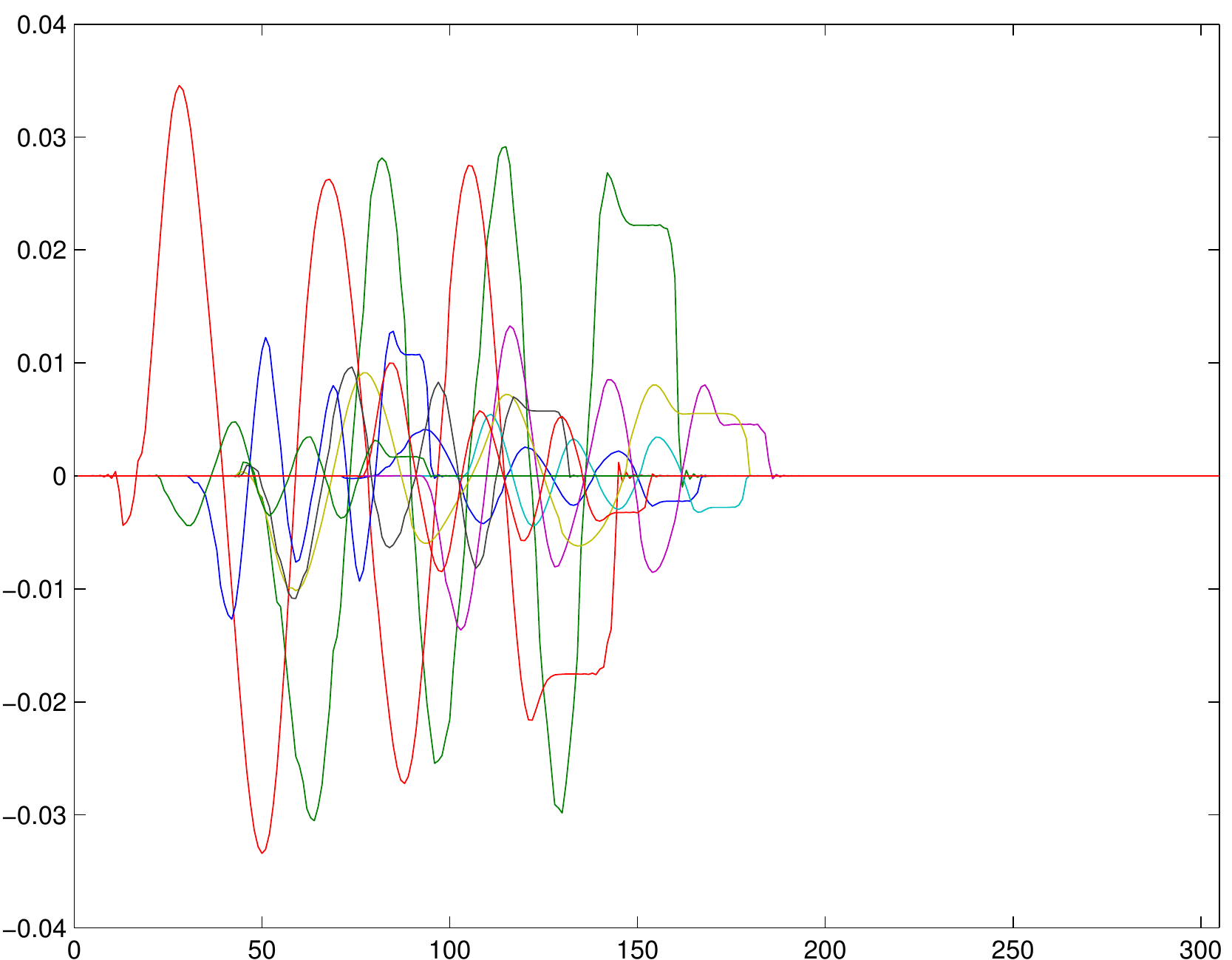}}
	\subfloat[Cluster 2 - Y]{
	\includegraphics[width=0.27\linewidth]{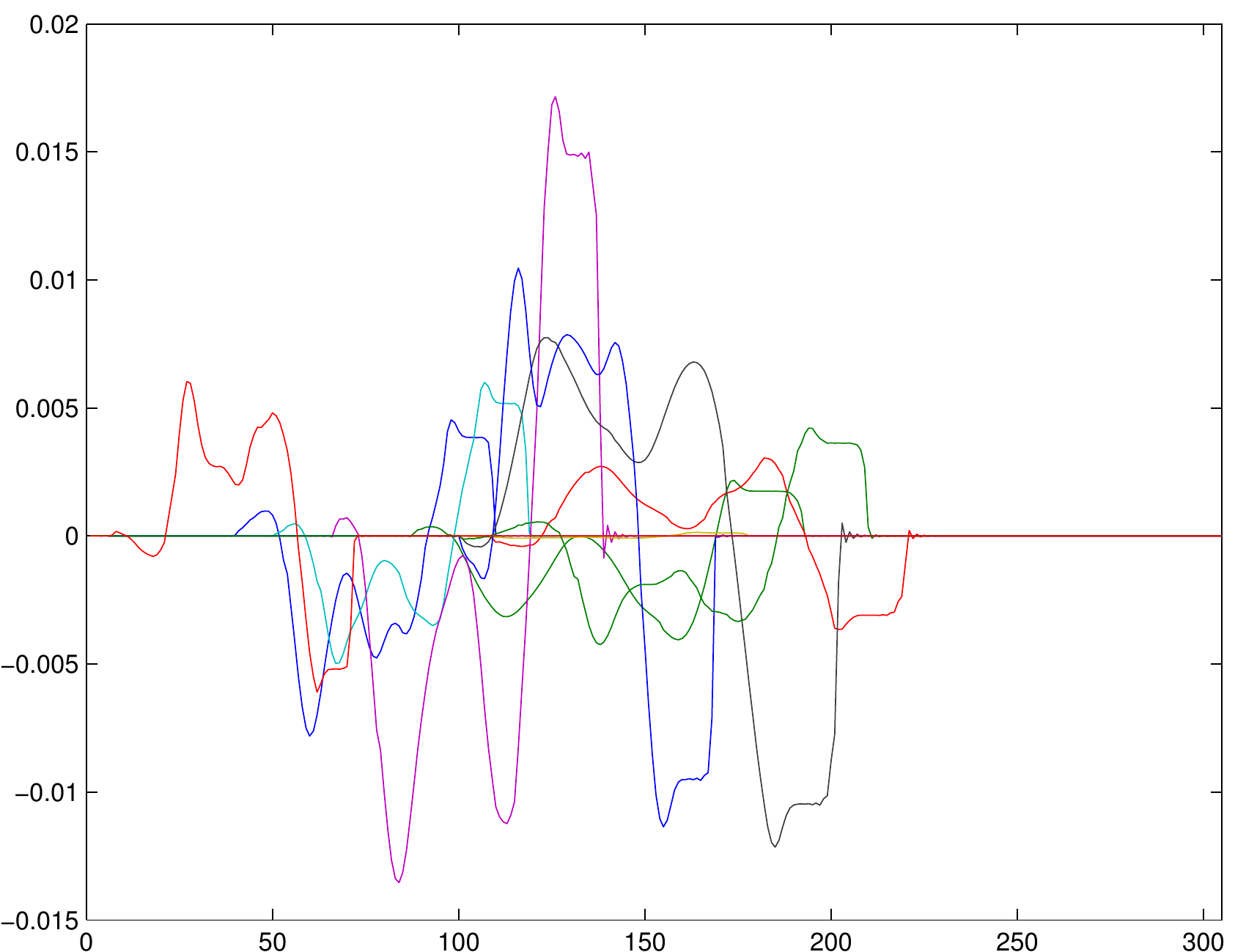}}
	\subfloat[Cluster 3 - Y]{
	\includegraphics[width=0.25\linewidth]{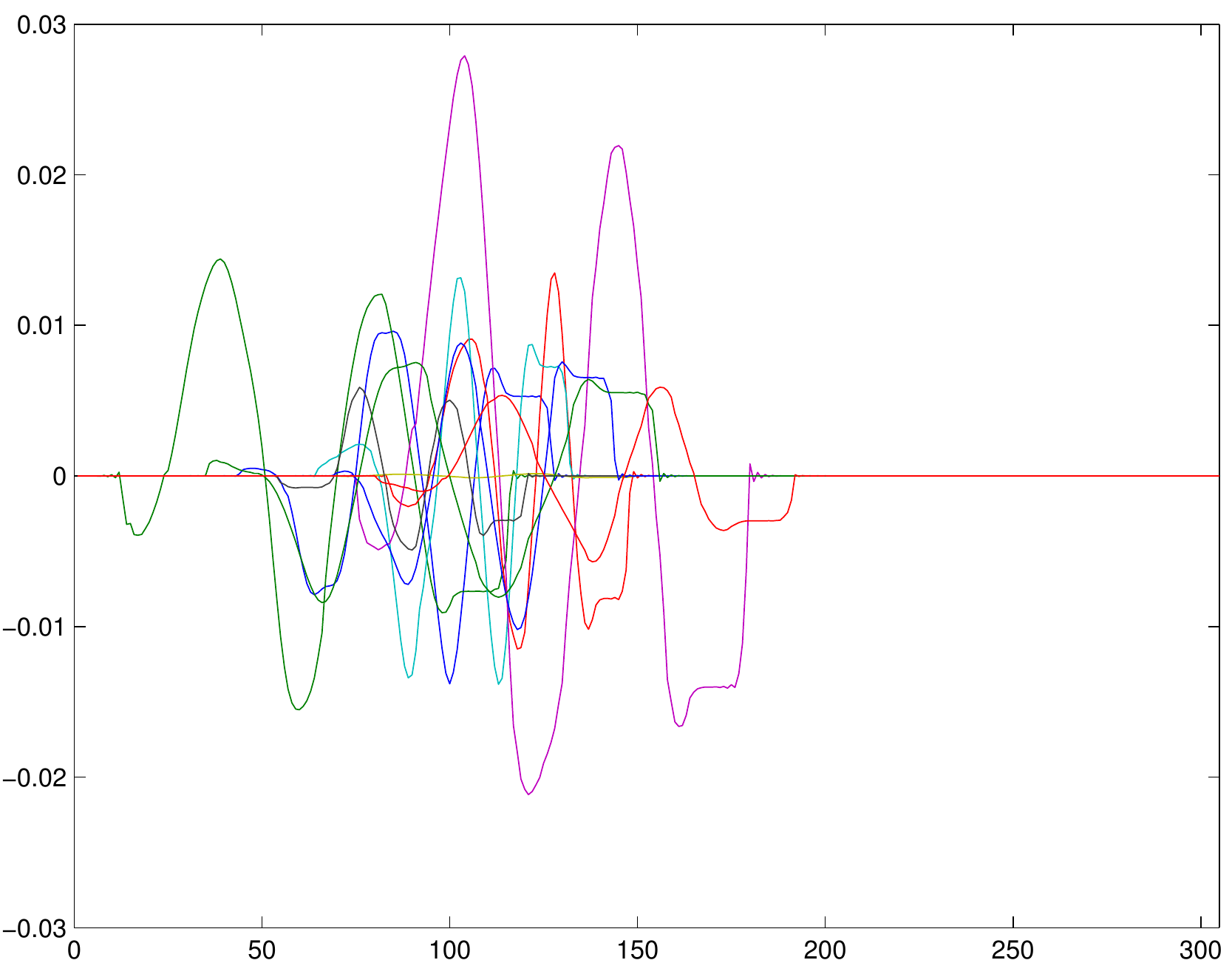}}
\caption{Example plots of curves used in the Character Classification Experiment. Each cluster consists of randomly selected characters from each class that are then subject to a combination of shifting, warping, stretching or shrinking and scaling. The top row shows the curves from the pen tip velocity in the X direction over time and the bottom row shows the same but for the Y direction.}
\label{fig_char_data}
\end{figure*}

In this experiment a collection of handwritten English characters were used to evaluate performance on a real world data set. The dataset consists of pen position data collected by a digitisation tablet at 200Hz, which is then converted to horizontal and vertical velocities \cite{williams2006extracting, williams2008modelling}. These 2-D trajectory curves are normalised such that the mean of each curve is close to zero. See Figure \ref{fig_char_examples} for some examples of this data. Figure \ref{fig_char_data} shows the example plots of curves used in the character classification experiment.

To evaluate performance twenty characters were randomly selected from three character classes. The data as originally released has been carefully produced and processed so that trajectories for each characters are extremely similar. Far more so than is realistic. For example the start time for each character has been aligned furthermore the writing speed, character size and variance in velocity over time is extremely consistent. Therefore to make the data more realistic we randomly globally shift each character so that their start times vary. Furthermore we randomly globally stretch and shrink each trajectory to account for different writing speeds, we also scale the trajectories by applying constant factors to account for character size and we lastly perform local warping (as done in the semi-synthetic experiment) to account for variance in speed over time. 

\begin{figure}[]
\centering
	\subfloat[LRR]{
	\includegraphics[width=1\linewidth]{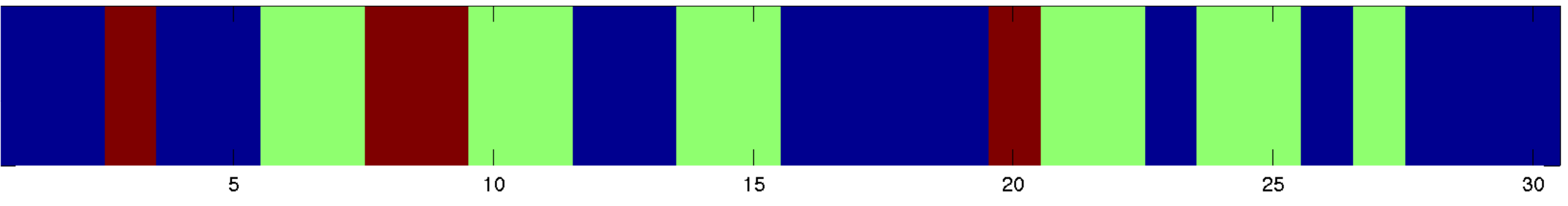}}\\
	\subfloat[Curve LRR]{
	\includegraphics[width=1\linewidth]{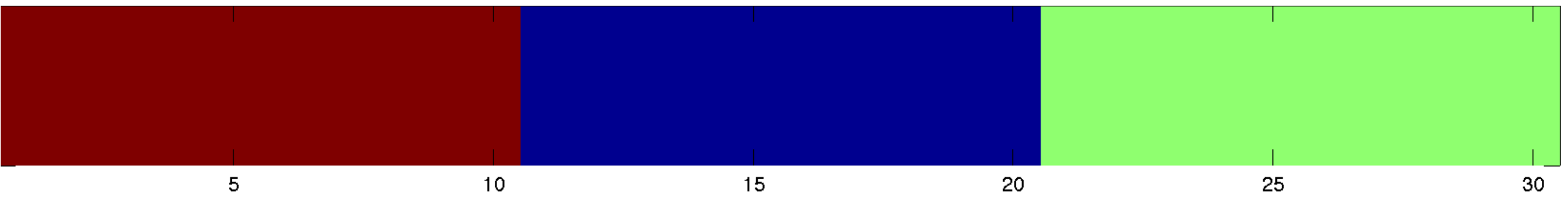}}
	
\caption{The segmentation results from the data in Figure \ref{fig_char_data}.}
\label{fig_char_clusters}
\end{figure}

Since the data consists of multidimensional curves the X and Y trajectory curves were concatenated to form data usable for conventional LRR since it can only handle vectors. Results can be found in Table \ref{table_char_results} and Figure \ref{fig_char_clusters}. Once again, the cLRR clearly outperforms LRR in all metrics. Furthermore cLRR shows excellent performance with a median accuracy of over $90\%$ on an extremely challenging dataset.

\begin{table}
\centering

\begin{tabular}{c c c c c}
\hline
 & Mean & Median & Min & Max \\
\hline

LRR			& 52.33\%			& 51.67\%		& 43.33\%		& 63.33\% \\
CurveLRR	& \bf 86.33\%		& \bf 91.67\%		& \bf 70\%		& \bf 100\%
	
\end{tabular}
\caption{Character Classification Results}
\label{table_char_results}
\end{table}

\section{Conclusion}\label{Sec:5}
In this paper, we extended the conventional LRR model on Euclidean space to a new LRR model for the manifold of open curves. The new LRR formulation is based on the tangent space approximation to the manifold so that the classic data self expressive can be well preserved for the manifold of curves at relevant high accuracy.  The resulting optimization problem can be solved using the LADMAP technique and algorithm convergence and complexity were presented. Finally we tested the new model by conducting experiments on synthetic, semi-synthetic and real world data, and the experimental results show the outstanding performance against the conventional LRR. Our next work is further extended the LRR model to the manifold of general closed curves.  

\section*{Acknowledgments}
Funding information hidden for the review process.

{\small
\bibliographystyle{ieee}
%\bibliography{references}

\begin{thebibliography}{10}\itemsep=-1pt

\bibitem{AbsilMahonySepulchre2008}
P.-A. Absil, R.~Mahony, and R.~Sepulchre.
\newblock {\em Optimization algorithms on matrix manifolds}.
\newblock Princeton University Press, 2008.

\bibitem{BahadoriKaleFanLiu2015}
M.~T. Bahadori, D.~Kale, Y.~Fan, and Y.~Liu.
\newblock Functional subspace clustering with application to time series.
\newblock In {\em Proceedings of The 32nd International Conference on Machine
  Learning}, pages 228--237, 2015.

\bibitem{Bishop2006}
C.~Bishop.
\newblock {\em Pattern Recognition and Machine Learning}.
\newblock Information Science and Statistics. Springer, 2006.

\bibitem{CaiCandesShen2008}
J.~F. Cai, E.~J. Cand\`{e}s, and Z.~Shen.
\newblock A singular value thresholding algorithm for matrix completion.
\newblock {\em SIAM J. on Optimization}, 20(4):1956--1982, 2008.

\bibitem{CandesLiMaWright2010}
E.~J. Cand\`{e}s, X.~Li, Y.~Ma, and J.~Wright.
\newblock Robust principal component analysis?
\newblock Submitted for publication, Stanford University, 2010.
\newblock \url{http://www-stat.stanford.edu/~candes/papers/RobustPCA.pdf}.

\bibitem{ElhamifarVidal2013}
E.~Elhamifar and R.~Vidal.
\newblock Sparse subspace clustering: {A}lgorithm, theory, and applications.
\newblock {\em IEEE Transactions on Pattern Analysis and Machine Intelligence},
  35(11):2765--2781, 2013.

\bibitem{FerratyRomain2011}
F.~Ferraty and Y.~Romain, editors.
\newblock {\em The Oxford Handbook of Functional Data Analysis}.
\newblock Oxford University Press, 2011.

\bibitem{FuGaoHongTien2015}
Y.~Fu, J.~Gao, X.~Hong, and D.~Tien.
\newblock Low rank representation on {R}iemannian manifold of symmetrical
  positive definite matrices.
\newblock In {\em SIAM Conferences on Data Mining (SDM)}, pages 316--324, 2015.

\bibitem{JoshiKlassenSrivastavaJermyn2007}
S.~H. Joshi, E.~Klassen, A.~Srivastava, and I.~Jermyn.
\newblock A novel representation for {R}iemannian analysis of elastic curves in
  $r^n$.
\newblock In {\em IEEE Conference on Computer Vision and Pattern Recognition},
  pages 1--7, 2007.

\bibitem{LinLiuLi2015}
Z.~Lin, R.~Liu, and H.~Li.
\newblock Linearized alternating direction method with parallel splitting and
  adaptive penalty for separable convex programs in machine learning.
\newblock {\em Machine Learning}, 99:287--325, 2015.

\bibitem{LinLiuSu2011}
Z.~Lin, R.~Liu, and Z.~Su.
\newblock Linearized alternating direction method with adaptive penalty for low
  rank representation.
\newblock In {\em Proceedings of NIPS}, 2011.

\bibitem{LiuLinYanSunYuMa2013}
G.~Liu, Z.~Lin, S.~Yan, J.~Sun, Y.~Yu, and Y.~Ma.
\newblock Robust recovery of subspace structures by low-rank representation.
\newblock {\em IEEE Transactions on Pattern Analysis and Machine Intelligence},
  35(1):171--184, 2013.

\bibitem{Mueller2011}
H.-G. M\"{u}ller.
\newblock {\em International Encyclopedia of Statistical Science}, chapter
  Functional data analysis, pages 554--555.
\newblock Springer, 2011.

\bibitem{parikh2013proximal}
N.~Parikh and S.~Boyd.
\newblock Proximal algorithms.
\newblock {\em Foundations and Trends in Optimization}, 1(3):123--231, 2013.

\bibitem{PetitjeanForestierWebbNicholsonChenKeogh2014}
F.~Petitjean, G.~Forestier, G.~I. Webb, A.~E. Nicholson, Y.~Chen, and E.~Keogh.
\newblock Dynamic time warping averaging of time series allows faster and more
  accurate classification. in icdm, 2014.
\newblock In {\em International Conference on Data Mining}, 2014.

\bibitem{Rakthanmanon2013}
T.~Rakthanmanon.
\newblock Addressing big data time series: Mining trillions of time series
  subsequences under dynamic time warping.
\newblock {\em ACM Transactions on Knowledge Discovery from Data}, 7(3):1--31,
  2013.

\bibitem{RamsaySilverman2005}
J.~Ramsay and B.~W. Silverman.
\newblock {\em Functional Data Analysis}.
\newblock Springer Series in Statistics. Springer, 2005.

\bibitem{Robinson2012}
D.~Robinson.
\newblock {\em Functional analysis and partial matching in the square root
  velocity framework}.
\newblock PhD thesis, Florida State University, 2012.

\bibitem{ShiMalik2000}
J.~Shi and J.~Malik.
\newblock Normalized cuts and image segmentation.
\newblock {\em IEEE Transactions on Pattern Analysis and Machine Intelligence},
  22:888--905, 2000.

\bibitem{SrivastavaShantanuJermyn2011}
A.~Srivastava, E.~Klassen, S.~H. Joshi, and I.~H. Jermyn.
\newblock Shape analysis of elastic curves in {E}uclidean spaces.
\newblock {\em IEEE Transactionson Pattern Analysis and Machine Intelligence},
  33(7):1415--1428, 2011.

\bibitem{SrivastavaWuKurtekKlassenMarron2011}
A.~Srivastava, W.~Wu, S.~Kurtek, E.~Klassen, and J.~S. Marron.
\newblock Registration of functional data using {Fisher-Rao} metric.
\newblock {\em varXiv:1103.3817}, 2011.

\bibitem{SuSrivastava2014}
J.~Su and A.~Srivastava.
\newblock Rate-invariant analysus of trajectories on {Riemannian} manifolds
  with application in visual speech recognition.
\newblock In {\em Proceedings of International Conference on Computer Vision
  and Pattern Recognition}, 2014.

\bibitem{SuSrivastavaHuffer2013}
J.~Su, A.~Srivastava, and F.~W. Huffer.
\newblock Detection, classification and estimation of individual shapes in {2D}
  and {3D} point clouds.
\newblock {\em Computational Statistics \& Data Analysis}, 58:227--241, 2013.

\bibitem{TuckerWuSrivastava2013}
J.~D. Tucker, W.~Wu, and A.~Srivastava.
\newblock Generative models for functional data using phase and amplitude
  separation.
\newblock {\em Computational Statistics and Data Analysis}, 61:50--66, 2013.

\bibitem{WangHuGaoSunYin2015}
B.~Wang, Y.~Hu, J.~Gao, Y.~Sun, and B.~Yin.
\newblock Low rank representation on grassmann manifolds: An extrinsic
  perspective.
\newblock {\em arXiv:1301.3529}, 1:1--9, 2015.

\bibitem{williams2008modelling}
B.~Williams, M.~Toussaint, and A.~J. Storkey.
\newblock Modelling motion primitives and their timing in biologically executed
  movements.
\newblock In {\em Advances in neural information processing systems}, pages
  1609--1616, 2008.

\bibitem{williams2006extracting}
B.~H. Williams, M.~Toussaint, and A.~J. Storkey.
\newblock {\em Extracting motion primitives from natural handwriting data}.
\newblock Springer, 2006.

\bibitem{WuGaneshShiMatsushitaWangMa2012}
L.~Wu, A.~Ganesh, B.~Shi, Y.~Matsushita, Y.~Wang, and Y.~Ma.
\newblock Convex optimization based low-rank matrix completion and recovery for
  photometric stereo and factor classification.
\newblock {\em IEEE Transactions on Pattern Analysis and Machine Intelligence},
  XX:XXX--XXX, August 2012.

\bibitem{YinGaoGuo2015}
M.~Yin, J.~Gao, and Y.~Guo.
\newblock A nonlinear low-rank representation on {S}tiefel manifold.
\newblock {\em Electronics Letters}, 51(10):749--751, 2015.

\bibitem{YinGaoLinShiGuo2015}
M.~Yin, J.~Gao, Z.~Lin, Q.~Shi, and Y.~Guo.
\newblock Graph dual regularized low-rank matrix approximation for data
  representation.
\newblock {\em IEEE Transactions on Image Processing}, 24(12):4918--4933, 2015.

\bibitem{ZhangSuKlassenLeSrivastava2015}
Z.~Zhang, J.~Su, E.~Klassen, H.~Le, and A.~Srivastava.
\newblock Video-based action recognition using rate-invariant analysis of
  covariance trajectories.
\newblock {\em arXiv:1503.06699v1}, 1, 2015.

\end{thebibliography}

}

\end{document}